\documentclass[review]{elsarticle}
\usepackage{setspace}
\usepackage{lineno,hyperref}
\modulolinenumbers[5]
\usepackage{float}
\usepackage{enumitem}
\newlist{steps}{enumerate}{1}
\setlist[steps, 1]{label = Step \arabic*:}
\usepackage{geometry}
\usepackage{algorithm}
\usepackage{algpseudocode}
\usepackage{pdfpages}
\usepackage{rotating}
\journal{Mechanism and Machine Theory}

\begin{document}
\begin{singlespace}
\begin{frontmatter}

\title{Degrees of Freedom Analysis of Mechanisms using a Novel Zebra Crossing Method}

\author{Rajashekhar V S \fnref{Corresponding author}}
\address{Project Assistant, Aerospace Engineering Department, Indian Institute of Science, Bangalore, Karnataka, India}
\fntext[myfootnote]{Corresponding author}
\ead{vsrajashekhar@gmail.com}

\author{Debasish Ghose}
\address{Professor, Aerospace Engineering Department, Indian Institute of Science, Bangalore, Karnataka, India}
\ead{dghose@iisc.ac.in}

%
%

\begin{abstract}
Mobility, which is a basic property, has to be analyzed to find the degrees of freedom of a mechanism. A novel and easy to implement method for calculation of degrees of freedom in a mechanism is proposed in this work. The mechanism is represented in a way that resembles a zebra crossing. An algorithm is proposed which is used to determine the mobility from the zebra crossing. This algorithm takes into account the number of patches between the black patches, the number of joints attached to the fixed link and the number of loops in the mechanism. A number of cases have been discussed, which fail to give the desired results using the widely used classical Kutzbach-Grubler formula but are amenable to the Zebra Crossing method proposed in this paper. We also show via an extensive comparison with other methods that the proposed methods is comparable to, and often better in terms of its implementability than, the best methods available in the literature.   
\end{abstract}


\end{frontmatter}

\linenumbers

\section{Introduction}
The degrees of freedom (DoF) or mobility calculation is an important property of mechanisms. It is defined by International Federation for the Promotion of Mechanism and Machine Science (IFToMM) as the number of independent coordinates needed to define the configuration of a kinematic chain or mechanism \cite{bgelsack1983terminology}. A literature review of the DoF works done in the past 150 years is given in \cite{gogu2005mobility}. The Chebychev-Grubler-Kutzbach criterion is widely used to determine the DoF of mechanisms. Many mechanisms do not obey the Chebychev-Grubler-Kutzbach criterion for calculating the DoF \cite{merlet2006parallel}. Quite often the classical quick formula fails to predict the correct number of DoF in modern mechanisms \cite{huang2009general}. Therefore, there is a need to find a quick formula to find the DoF of a mechanism that can be applied to modern multi-loop mechanisms, parallel manipulators, and other classical mechanisms which cannot otherwise be calculated using the existing quick methods.  

The past two decades have seen mobility formulas being proposed for a particular family of mechanisms. It can be observed that the ways to determine the mobility of a mechanism can be broadly classified into three methods as mentioned in \cite{muller2009generic}. The most commonly used method is the quick formulas that takes into account the number of links, joints, loops and other parameters of the mechanism. The second method is finding the rank of the Jacobian matrix which gives the DoF of the mechanism. The third method is the usage of screw theory and other mathematical techniques. They fail to calculate the DoF of modern mechanisms such as the Cartesian Parallel Manipulator. A few of the works done in the past two decades are summarized here. 

A formula \cite{yang2012general} for DoF of parallel mechanisms and multiloop spatial mechanisms using vector algebra and theory of sets has been proposed. In \cite{huang2009general} a method based on screw theory has been proposed to find the mobility of mechanisms. Lie algebra is used to determine the mobility criterion  of parallel platforms in \cite{rico2004more}. The mobility of spatial mechanisms is calculated by considering polyhedral model \cite{wampler2007new}. Using Jacobian matrices, mobility of parallel mechanisms can be determined \cite{yang2008simple}. The DoF for complex spatial mechanisms is calculated by  introducing a theory and the concept of configuration
degrees of freedom \cite{zhao2004theory}. Mobility of multi-loop mechanisms is found by a combination of bio-inspired cell division method and screw theories \cite{li2019cell}. A mobility condition is obtained using Groebner basis which takes into account the dimensional and positional parameters of the links \cite{rameau2015computing}. Similarly, the effect of dimensions of links in a mechanism on the mobility is discussed in \cite{rameau2016dimensional}. A constraint graph that allows representation of all the constraints which includes multiple joints is introduced in \cite{muller2017constraint} from which the mobility can be determined. Fast and accurate calculation of mobility of overconstraint planar mechanisms are given in \cite{zeng2016determination} by introducing new concepts and theories. The concept of generic mobility which takes into consideration the presence of link imperfection is introduced in \cite{muller2009generic}. The mobility in kinematic chains cannot be determined correctly using the velocity analysis but can be determined when combined with Lie algebra \cite{rico2009infinitesimal}. The dynamic models that were less used previously were used to establish the mobility criteria in multi-loop mechanisms \cite{talaba2015mechanical}. A method to find the DoF of the mechanism by using the Macaulay matrices at a given configuration is proposed \cite{wampler2011mechanism}. The deployable assembly mechanisms have modules that are identical. The concept of null space is used to find the mobility of such mechanisms \cite{yang2016mobility}. Identification of redundant and passive mobilities using the intersection of screw manifolds in parallel manipulators is proposed in \cite{bu2016mobility}. The complex ball mechanism is decomposed and analyzed using the screw system analysis \cite{dai2004mobility}. A method for analyzing limited degrees of freedom parallel manipulators, using geometric algebra, is proposed in \cite{li2016mobility}. Mobility of overconstrained parallel mechanisms using Grassmann-Cayley algebra is given in \cite{chai2017mobility}. The mobility of parallel manipulators can be calculated by analyzing the pattern of the transformation matrix \cite{chen2010mobility}. Screw theory is used to find the mobility of single-loop mechanisms \cite{milenkovic2010mobility}, parallel mechanisms \cite{wang2015mobility}, overconstrained parallel mechanisms \cite{dai2006mobility}, and spatial mechanisms \cite{zeng2015over}. The mobility of single-loop overconstrained mechanisms is calculated using Jacobian matrices \cite{rico2007calculating}. The mobility is calculated by finding the rank of position and orientation characteristic equation of single loop mechanisms in \cite{yang2008rank}.   

It can be observed that many works have focused on screw theory \cite{huang2009general,bu2016mobility,dai2004mobility,milenkovic2010mobility,wang2015mobility,dai2006mobility,zeng2015over} and on finding the rank of Jacobian matrices \cite{yang2008simple,rico2007calculating} as a base for determing the DoF of mechanisms, while the quick methods have not received much attention in the past two decades. It can be noted that works have emerged taking into consideration the dimensions of the links \cite{rameau2015computing,rameau2016dimensional} for finding the DoF. The other areas of mathematics have been explored to find the DoF including vector algebra and theory of sets \cite{yang2012general}, Lie algebra \cite{rico2004more,rico2009infinitesimal} and Grassmann-Cayley algebra \cite{chai2017mobility}. Bio-inspired methods have also started to emerge \cite{li2019cell}. 

The quick methods fail in certain cases where the mechanisms have loops. This failure of identifying the right degree of freedom is due to the inability to find the kinematically dependant loops in the mechanisms. A detailed explanation for this is given in \cite{gogu2005mobility}. Hence a detailed analysis is required for identifying the right degree of freedom. There are cases \cite{freudenstein1975degree} where these methods also fail to give the right degree of freedom. The Zebra Crossing method (a quick method) proposed in this work provides the right degree of freedom even for mechanisms that get the wrong results using the formulas available in the literature. 

\section{The Zebra Crossing Algorithm}
The zebra crossing algorithm helps in finding the degrees of freedom for a mechanism. The given mechanism is converted into a zebra crossing diagram consisting of black, white and grey patches. The degrees of freedom is found based on the number of loops in the mechanism and on counting the number of black, white, and grey patches.  
 
\subsection{Conversion to Zebra Crossing diagram}
\begin{figure}[H]
\begin{center}
\includegraphics[scale=0.50]{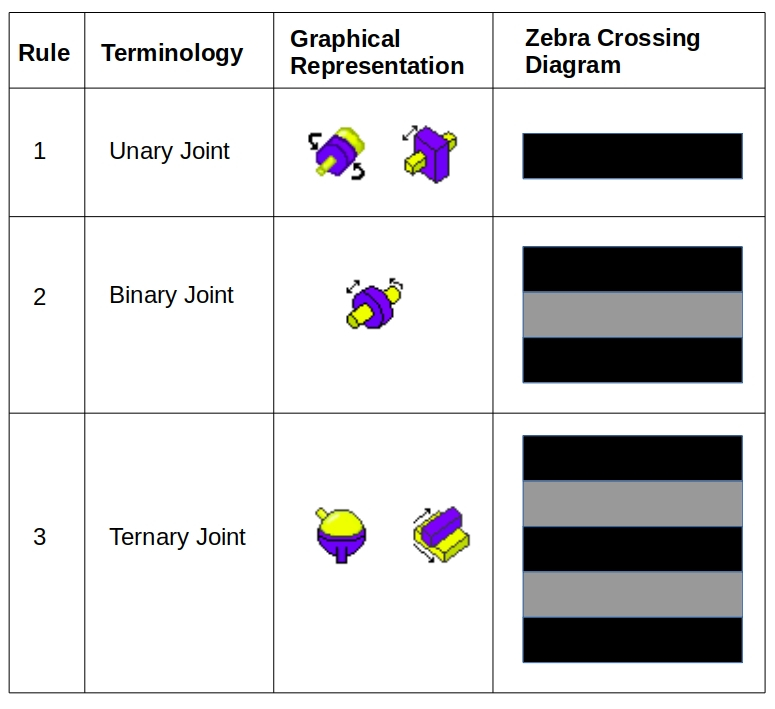} 
\caption{Representation of joints for the Zebra Crossing diagram}
\label{fig_joints}
\end{center}
\end{figure}

\begin{figure}[H]
\begin{center}
\includegraphics[scale=0.50]{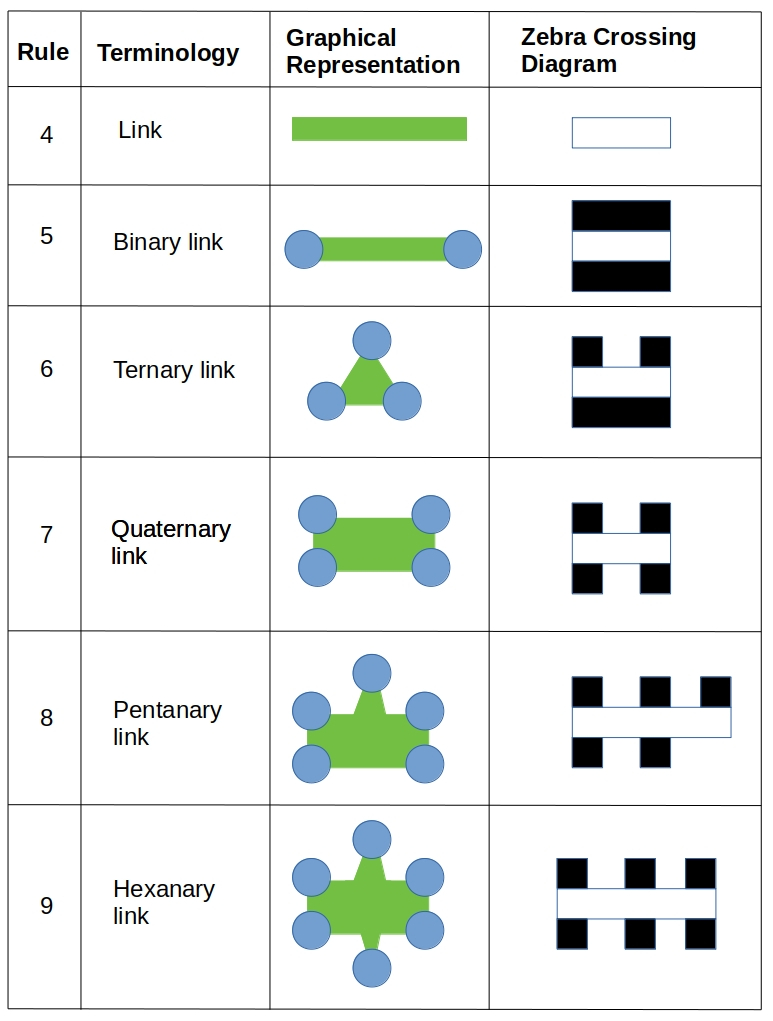} 
\caption{Representation of links for the Zebra Crossing diagram}
\label{fig_links}
\end{center}
\end{figure}

The given mechanism is first converted into a Zebra Crossing diagram. The various elements of the Zebra Crossing diagram are shown in Figure \ref{fig_joints} and \ref{fig_links}. In Figure \ref{fig_joints} we show how the unary, binary and ternary joints are represented in Zebra Crossing diagram. In Figure \ref{fig_links} we show how the links, binary, ternary and quaternary links are represented in Zebra Crossing diagram.  


\subsection{Rules for converting a mechanism diagram to Zebra Crossing diagram}
It is well known that the mechanism is a collection of links and joints. It can either be open loop or closed loop in nature. For open loop mechanisms the number of joints will be equal to the number of degrees of freedom. For closed loop mechanisms the number of degrees of freedom will always be less than the number of joints. The following are the nine rules that govern the conversion of the mechanism diagram to the Zebra Crossing diagram. This is shown in Figure \ref{fig_joints} and Figure \ref{fig_links}.
\textit{Rule 1}: It represents the unary joint which has one degree of freedom. It can be revolute joint or prismatic joint. In the zebra crossing diagram it is represented by a single black patch. 

\textit{Rule 2}: It represents the binary joint which has two degrees of freedom. It can be cylindrical joint or universal joint. In the zebra crossing diagram it is represented by a single grey patch in between two black patches.

\textit{Rule 3}: It represents the ternary joint which has three degrees of freedom. It can be spherical joint or planar joint. In the zebra crossing diagram it is represented by two grey patches in between three black patches.

\textit{Rule 4}: It represents the link of the mechanism. In the zebra crossing diagram it is represented by a white patch.

\textit{Rule 5}: It represents the binary link of the mechanism. This means that two joints can be attached to the link. In the zebra crossing diagram it is represented by a white patch in between two black patches.

\textit{Rule 6}: It represents the ternary link of the mechanism. This means that three joints can be attached to the link. In the zebra crossing diagram it is represented by a white patch in between three black patches.

\textit{Rule 7}: It represents the quaternary link of the mechanism. This means that four joints can be attached to the link. In the zebra crossing diagram it is represented by a white patch in between four black patches.

\textit{Rule 8}: It represents the pentanary link of the mechanism. This means that five joints can be attached to the link. In the zebra crossing diagram it is represented by a white patch surrounded by five black patches.

\textit{Rule 9}: It represents the hexanary link of the mechanism. This means that six joints can be attached to the link. In the zebra crossing diagram it is represented by a white patch surrounded by six black patches.

The conversion to the Zebra Crossing diagram ends once the last link or the white patch is reached. This white patch is the fixed link.

\subsection{Analogy between mechanisms, graphs and zebra crossing algorithm}
In \cite{tsai2000mechanism}, graph theory is used as a tool to represent mechanisms. A comparison is made here between the various parameters of a mechanism, graph theory, and the proposed Zebra Crossing method. The links in the mechanism is equivalent to the vertices of the graph and white patch in the zebra crossing. The joints in the mechanism is equivalent to the edges in the graph and black patch in the zebra crossing. The other parameters comparing the mechanism with the graph theory and zebra crossing are listed in Table \ref{tab_analogy}. 

\begin{table}[H]
\caption{A analogy between mechanisms, graphs, and zebra crossing method}
\label{tab_analogy}
\begin{tabular}{|l|l|l|}
\hline
\multicolumn{1}{|c|}{\textbf{Mechanism}}                                     & \multicolumn{1}{c|}{\textbf{Graph Theory}}                                  & \multicolumn{1}{c|}{\textbf{Zebra Crossing Method}}                                               \\ \hline
Links                                                                        & Vertices                                                                    & White patch                                                                                       \\ \hline
Joints                                                                       & Edges                                                                       & Black Patch                                                                                       \\ \hline
\begin{tabular}[c]{@{}l@{}}Number of links \\ having “i” joints\end{tabular} & \begin{tabular}[c]{@{}l@{}}Number of vertices \\ of degree “i”\end{tabular} & \begin{tabular}[c]{@{}l@{}}Number of white patches \\ connected to “i” black patches\end{tabular} \\ \hline
Number of joints on link “i”                                                 & Degree of vertex “i”                                                        & \begin{tabular}[c]{@{}l@{}}Number of black patches \\ connected to "i" white patches\end{tabular} \\ \hline
Loops=Joints - Links + 1                                                     & Loops=Edges - Vertices + 1                                                  & \begin{tabular}[c]{@{}l@{}}Loops=Total Black Patches - \\ Total White Patches + 1\end{tabular}    \\ \hline
Isomorphic mechanisms                                                        & Isomorphic graphs                                                           & Isomorphic zebra crossing                                                                         \\ \hline
\end{tabular}
\end{table}

In \cite{tsai2000mechanism}, although graph theory is used in enumeration of the mechanisms, their degrees of freedom is calculated using the Kutzbach-Grubler formula. It can be seen that the number of loops are calculated by forming similar formulas from links and joints in graph theory and zebra crossing algorithm.  
\subsection{The zebra crossing algorithm}
\begin{figure}[H]
\begin{center}
\includegraphics[scale=0.7]{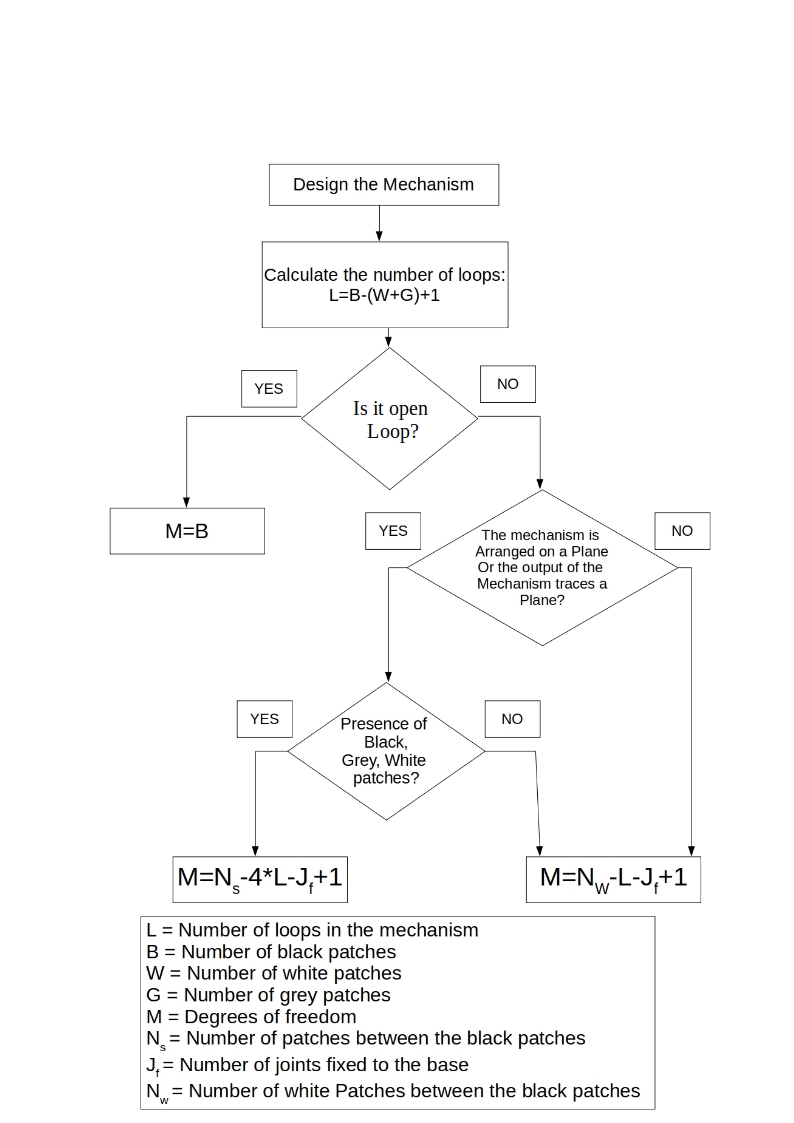} 
\caption{The zebra crossing algorithm}
\label{fig_algorithm}
\end{center}
\end{figure}

The zebra crossing algorithm is summarized in Figure \ref{fig_algorithm}. The algorithm is explained as follows.
\begin{steps}
\item Represent the mechanism by sketching clearly its links and joints.
\item Represent the mechanism by using zebra crossing diagram using Rules $1-9$ given above. 
\item Count the number of black, grey and white patches in the zebra crossing diagram. Use Equation \ref{equ_loop} to find the total number of loops in the mechanism. 
\begin{equation}
L = B-\left(W+G\right)+1
\label{equ_loop}
\end{equation} 
where,\\
$L$ = Number of loops in the mechanism\\
$B$ = Number of black patches\\
$W$ = Number of white patches\\
$G$ = Number of grey patches\\
\item If the number of loops are $0$, then it is a open loop mechanism. The degrees of freedom equal the number of black patches in the zebra crossing diagram.  
\begin{equation}
M = B
\label{equ_open_loop_dof}
\end{equation}
\item If the number of loops are more than $0$, then it is a closed loop mechanism. Find whether the mechanism can have a configuration where its links and joints lie on a plane.
\item If the mechanism can be placed on a plane in atleast one configuration and has black, grey, and white patches use Equation \ref{equ_loopplanar} below to find its degrees of freedom.
\begin{equation}
M = N_{s} - \left(4 \times L\right) - J_{f} + 1
\label{equ_loopplanar}
\end{equation}  
where,\\
$M$ = Degrees of freedom of the mechanism\\
$N_{s}$ = Number of patches between the black patches\\
$L$ = Number of loops in the mechanism\\
$J_{f}$ = Number of joints attached to the ground link\\
\item If the mechanism cannot be placed on a plane, that is the mechanism is spatial in configuration, and has black, grey and white patches or only black and grey patches, then use Equation \ref{equ_loopwhite} to find its degree of freedom. If it is a parallel manipulator, then the manipulator platform is treated as link. When counting the number of white patches, this white patch is assigned a number which is one less than the number of legs in the parallel manipulator.    
\begin{equation}
M = N_{w} - L - J_{f} + 1
\label{equ_loopwhite}
\end{equation}
where,\\
$M$ = Degrees of freedom of the mechanism\\
$N_{w}$ = Number of white patches between the black patches\\
$L$ = Number of loops in the mechanism\\
$J_{f}$ = Number of joints attached to the ground link\\
\end{steps}

By following the above steps, the degrees of freedom can be obtained. These steps are written in the form of an algorithm as follows. 
\begin{algorithm}
   \caption{Zebra Crossing Algorithm}
    \begin{algorithmic}[1]
     \State Design a mechanism based on the requirement and marks its links and joints clearly.
	  \State Convert it into the Zebra Crossing diagram. 
	  \State Calculate the number of loops using the formula: $L = B-\left(W+G\right)+1$
	  \If {L = 0}
	   \State $M = B$
	  \ElsIf {the mechanism can be arranged in a plane and has black, grey and white patches in the Zebra Crossing diagram}
	  		\State $M = N_{s} - \left(4 \times L\right) - J_{f} + 1$
	  		\ElsIf {the mechanism can be arranged in a plane and has only Black and White patches in the Zebra Crossing diagram}  
	  		\State $M = N_{w} - L - J_{f} + 1$
	  \Else 
	  \State the mechanism cannot be arranged in a plane
	   \State $M = N_{w} - L - J_{f} + 1$
	  \EndIf
\end{algorithmic}
\end{algorithm}

\section{Theoretical Foundation of the Zebra Crossing Algorithm}
\subsection{Four Lemmas}
\subsubsection{Lemma A: Patch Consideration}
By Rules 1-3, a joint with \textit{f} degrees of freedom is drawn as \textit{f} black patches and has \textit{(f - 1)} grey patches; by Rules 4-9 a link has one white patch. Summing over the diagram:

\begin{equation}
B = \Sigma F
\end{equation}

\begin{equation}
G = \Sigma ( f - 1)
\end{equation}

\begin{equation}
G = B - j
\end{equation}

which when rearranged gives,
\begin{equation}
j = B - G
\end{equation}

So the black patches count joint freedoms, the black-patch groups count joints, and the white patches count links, 

\begin{equation}
W = N
\end{equation}

\subsubsection{Lemma B: Circuit rank}
The mechanism is considered as a connected graph (links = vertices, joints = edges). A spanning tree uses N - 1 edges and every one of the remaining j - (N - 1) edges closes exactly one independent loop. 

\begin{equation}
L = j - N + 1 
\end{equation}

\begin{equation}
L = (B - G) - W + 1 
\end{equation}

\begin{equation}
L = B - (W + G) + 1
\end{equation}

This is same as the equation \ref{equ_loop} to find the total number of loops in the mechanism.

\subsubsection{Lemma C: Loop form of the CGK criterion}
The configuration of the mechanism is described by $\Sigma$f joint variables. If all constraints are independent, the mobility is 

\begin{equation}
M = \Sigma f - \lambda L
\end{equation} 

This is equivalent to the classical CGK formula

\begin{equation}
M = \lambda (N - 1 - j) + \Sigma f
\end{equation}

\subsubsection{Lemma D: The structural counting Lemma}
\paragraph{Opening the loop at a ground joint}
If there are $J_f$ joints attached to the ground, the fixed link's white patch is connected to only one of them in the strip representation.

\begin{equation}
L = J_f - 1
\end{equation}

\paragraph{Opening the loop by cutting a moving link:}
Consider a 4-bar mechanism where the white patch is left at the end of a branch and is no longer between black patches. Here one white patch is lost per cut (c cuts). 

\paragraph{Keep the loop closed but merging duplicate patches:}
Instead of cutting the loop, sometimes the diagram is allowed to remain closed. In this case the same physical link can appear through two branches. Instead of counting them as separate white patches, they are counted as same. This is called as a merge (m). 

\begin{equation}
\label{equ:lemma4_equ1}
L = (J_f - 1) + c + m
\end{equation}

\begin{equation}
\label{equ:lemma4_equ2}
N_w = (N - 1) - (c + m)
\end{equation}

Rearranging Equation \ref{equ:lemma4_equ1},

\begin{equation}
c + m = L - (J_f - 1)
\end{equation}

or equivalently,

\begin{equation}
c + m = L - J_f + 1.
\end{equation}

Substituting this result into Equation \ref{equ:lemma4_equ2},

\begin{equation}
N_w = (N - 1) - (L - J_f + 1).
\end{equation}

From Lemma B, $(N - 1 = j - L)$, we get, 

\begin{equation}
\label{equ_lemma_d}
N_w  = j - 2L + J_f  - 1
\end{equation}

%

Finally, every grey patch lies between two black patches by construction (Rules 2-3), so
\begin{equation}
\label{equ_between_all}
N_s  = N_w  + G 
\end{equation}

\subsection{Derivation of Equation \ref{equ_loopplanar}}
We know that the mechanism can be placed in a plane in at least one configuration, and its diagram contains black, grey, and white patches. Grey patches signify multi-DoF joints. The motion parameter is $\lambda = 6$.

We know that Loop-form mobility (Lemma C) with $\lambda = 6$:

\begin{equation}
\label{requ_grey_white_black}
M = \Sigma f - 6L
\end{equation}

From Lemma A ($\Sigma f = B$),

\begin{equation}
M = B - 6L
\end{equation}

Rearranging equation \ref{equ_lemma_d}, we get

\begin{equation}
j = N_w + 2L - J_f + 1
\end{equation}

By including Lemma A $( B = j + G )$ and then equation \ref{equ_between_all},

\begin{equation}
B = j + G
\end{equation}

\begin{equation}
B = N_w + 2L - J_f + 1 + G
\end{equation}

\begin{equation}
B = N_s + 2L - J_f + 1
\end{equation}

Substitute into equation \ref{requ_grey_white_black}, we get:

\begin{equation}
M = ( N_s  + 2L - J_f  + 1 ) - 6L
\end{equation}

\begin{equation}
M = N_s  - 4L - J_f  + 1
\end{equation} 

This is equation \ref{equ_loopplanar}.

\subsection{Derivation of Equation \ref{equ_loopwhite} - Application 1}
Here the mechanism can be arranged in a plane, and the zebra crossing diagram of the mechanism has only black and white patches. The absence of grey patches means that every joint has one DoF (revolute or prismatic, Rule 1). Hence $\Sigma f = j$, the motion is planar, and $\lambda = 3$.

From Lemma C with $\lambda = 3$ and $\Sigma f = j$, we have

\begin{equation}
M = j - 3L
\end{equation} 

From Lemma D,

\begin{equation}
j = N_w  + 2L - J_f  + 1
\end{equation}

By combining the above equations, we get: 

\begin{equation}
M = N_w  - L - J_f  + 1 
\end{equation}

Here the loop coefficient is (3 - 2 = 1): three planar closure constraints per loop minus the same 2 loops. 

It is to be observed that Equations \ref{equ_loopplanar} and \ref{equ_loopwhite} are one formula, where 
\begin{equation}
M = (patch count) - (\lambda - 2)L - J_f + 1
\end{equation}

They are evaluated at $\lambda$ = 6 and $\lambda$ = 3.

\subsection{Derivation of Equation \ref{equ_loopwhite} - Application 2}
Here, spatial parallel manipulators with a fixed base, one moving platform, and $n$ legs are considered. There is physically only one platform. In this case, while drawing the zebra crossing diagram, the fixed base is not drawn. Each leg starts at its base joint, and all the legs terminate at the same white patch representing the moving platform. Here, the entire mechanism is treated as one continuous chain:

\[
\mathrm{Leg\ 1} \rightarrow \mathrm{platform} \rightarrow
\mathrm{Leg\ 2} \rightarrow \mathrm{platform} \rightarrow
\mathrm{Leg\ 3}
\]

This means that, if there are $n=3$ legs, then there are $n-1$, which is 2; therefore the platform contributes for 2 white patches. When the whole manipulator is treated as one continuous chain, the platform is traversed (n - 1) times. This is once between each consecutive pair of legs. Hence the white patch between black patches ($N_w$) must be counted as (n - 1) for the moving platform. 

The multiple white-patch representation does not imply multiple physical platforms; it is a topological unfolding of a single platform into a continuous traversal representation used only for representing branch connections in the counting procedure. In this each traversal between consecutive legs represents one independent branch connection.

\begin{equation}
M = N_w  - L - J_f  + 1 
\end{equation}

\subsection{Why the Zebra Crossing Algorithm succeeds where Chebychev-Grübler-Kutzbach method fails?}

The counting of a multi-DoF joint as one unit discards its passive freedoms. Moreover, the per-leg count is insensitive to the overconstraints that cause the Chebychev--Grübler--Kutzbach method to fail for Cartesian manipulators such as the Orthoglide, STAR, and DELTA parallel manipulators. All three results are ultimately based on the loop-form Chebychev--Grübler--Kutzbach criterion,

\begin{equation}
M = \Sigma f - \lambda L
\end{equation}

which is re-expressed through zebra crossing diagram counting. The zebra crossing conventions introduce two kinematically meaningful features beyond the Chebychev--Grübler--Kutzbach method. First, grey-patch grouping absorbs passive joint freedoms. Second, the merge-versus-keep treatment preserves the structural identity of repeated physical links when the mechanism is unfolded into the Zebra Crossing representation. This is why the method returns correct results for overconstrained and modern parallel mechanisms for which classical shortcut formulations, such as the Chebychev--Grübler--Kutzbach method, fail.

\section{Examples}
In this section we consider several examples that have been solved or unsolved in the literature to show that our method works in all of them while others do not. 
\subsection{Open loop mechanisms}
\subsubsection{One DOF helical joint mechanism}
\begin{figure}[H]
\begin{center}
\includegraphics[scale=0.4]{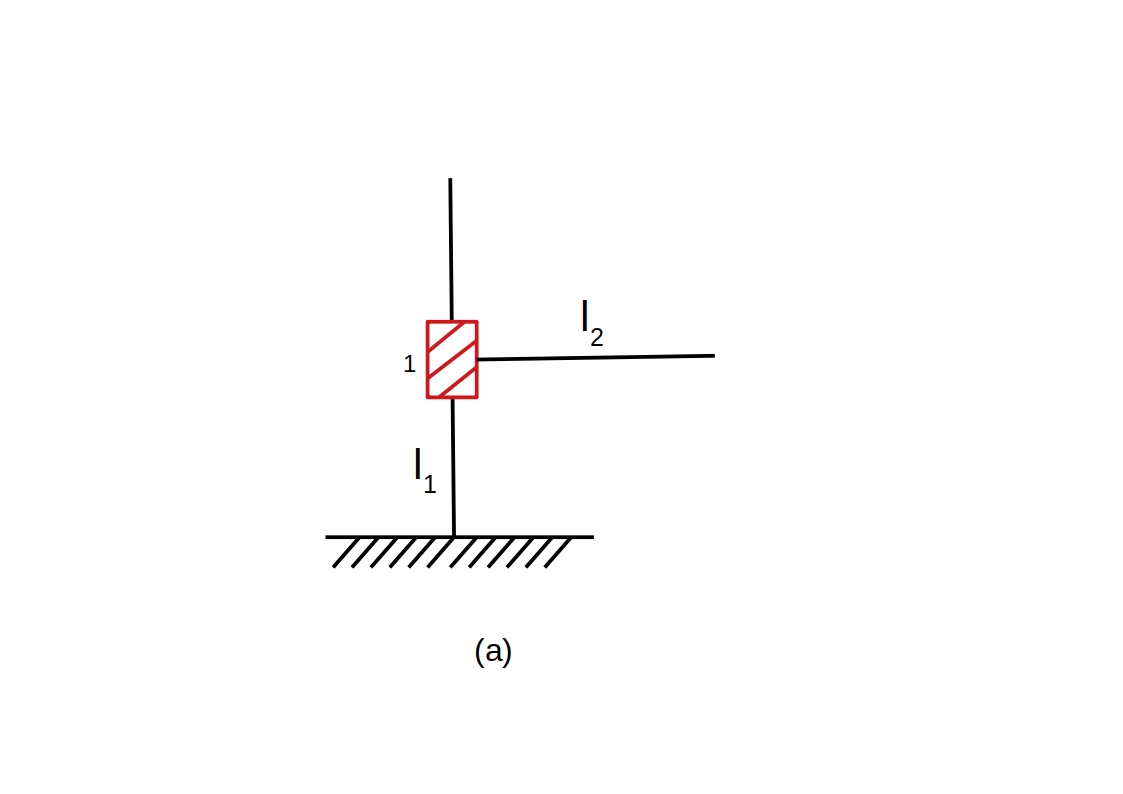} 
\includegraphics[scale=0.4]{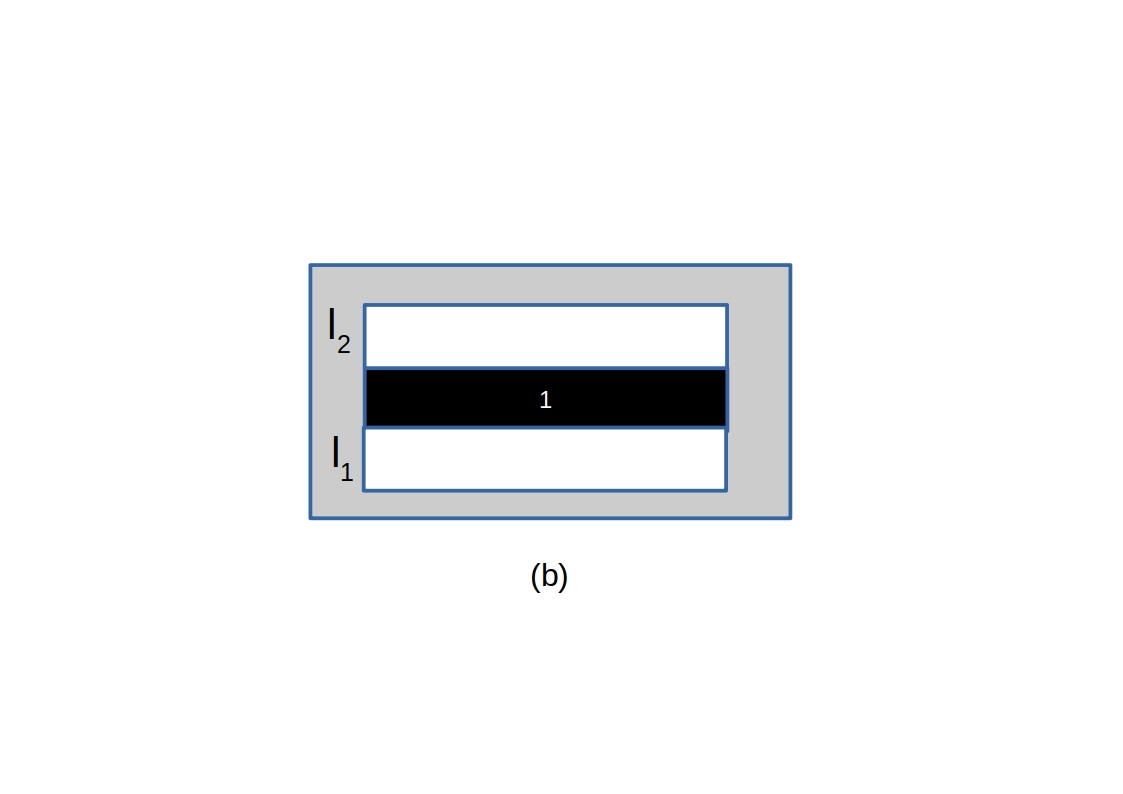} 
\caption{The zebra crossing algorithm applied to the one DOF helical joint mechanism (a) The one DOF helical joint mechanism (b) The zebra crossing diagram of the mechanism}
\label{fig_8}
\end{center}
\end{figure}
One DOF helical joint mechanism is a simple mechanism with two links connected by a helical joint. This is shown in Figure \ref{fig_8} (a). The Zebra crossing diagram of the mechanism is drawn as follows. The \textit{Rule 4} is applied which represents the fixed link. This is followed by \textit{Rule 1} and \textit{Rule 4}. This completes the Zebra crossing diagram and is shown in Figure \ref{fig_8} (b).
   
From the Zebra Crossing diagram, the number of loops (L) in the mechanism are calculated using the Equation \ref{equ_loop}. There is $1$ black patch and $2$ white patches in the zebra crossing diagram. Therefore, 
\begin{equation}
L=1-2+1=0
\end{equation}
The number of loops in the mechanism is found to be $0$. Hence it is an open loop mechanism. 
Calculating the degrees of freedom (M) from Equation \ref{equ_open_loop_dof},
\begin{equation}
M=1
\end{equation}
Therefore, the mechanism has $1$ degree of freedom. 
The above steps can be followed to find the degrees of freedom in open loop mechanisms. 
\subsection{Closed loop mechanisms}
\subsubsection{Planar mechanisms}
\paragraph{Four-bar mechanism}
\begin{figure}[H]
\begin{center}
\includegraphics[scale=0.4]{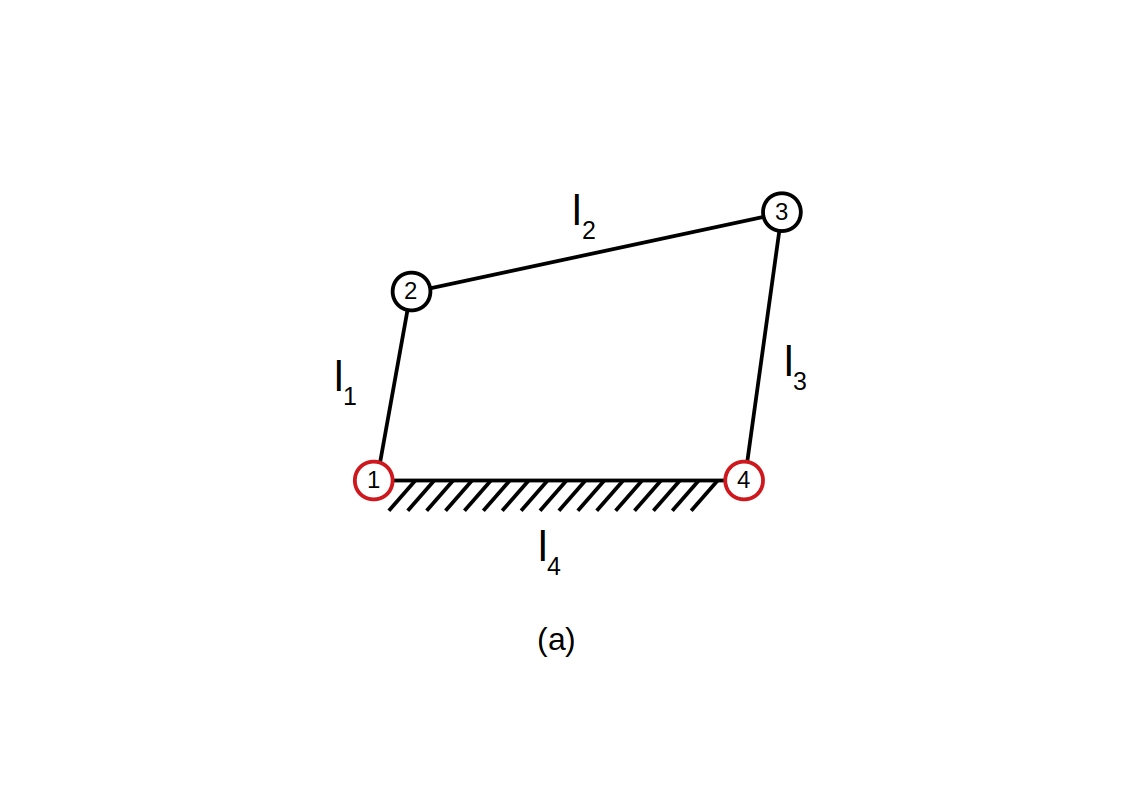} 
\includegraphics[scale=0.4]{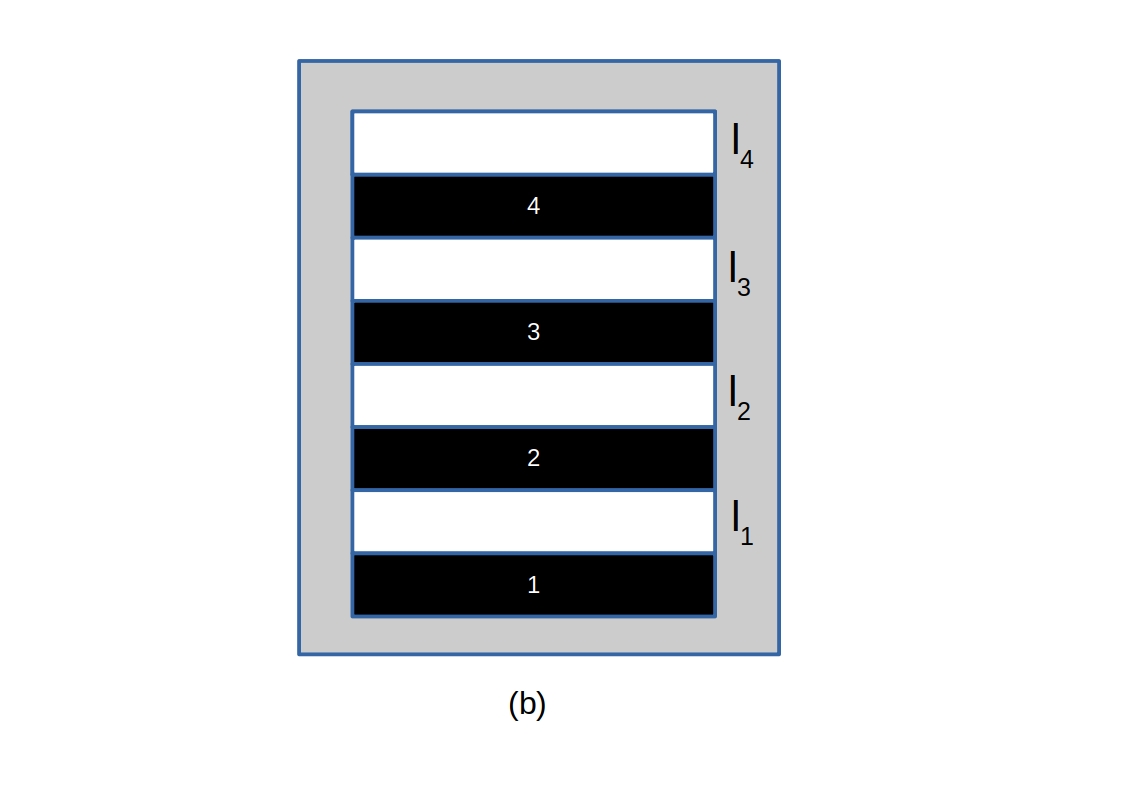}
\caption{The zebra crossing algorithm applied to the four bar mechanism (a) The four bar mechanism (b) The zebra crossing diagram of the mechanism}
\label{fig_1}
\end{center}
\end{figure}
The four bar mechanism is shown in Figure \ref{fig_1} (a). The Zebra crossing diagram is drawn by following the \textit{Rule 1} followed by \textit{Rule 4}. These two rules are repeated till the link \textit{$l_4$} is reached.

There are $4$ black patches and $4$ white patches in the Zebra crossing diagram. Applying Equation \ref{equ_loop}, the number of loops (L) in the mechanism are 
\begin{equation}
L=4-4+1=1
\end{equation}
The number of white patches between the black patches $N_w$ are $3$. The number of loops $L$ is $1$. The number of joints attached to the ground link $J_f$ are $2$. The degrees of freedom (M) is calculated using the Equation \ref{equ_loopwhite},
\begin{equation}
M=3-1-2+1=1
\end{equation}
Therefore, the mechanism has $1$ degree of freedom. 

\paragraph{Staircase mechanism}
\begin{figure}[H]
\begin{center}
\includegraphics[scale=0.4]{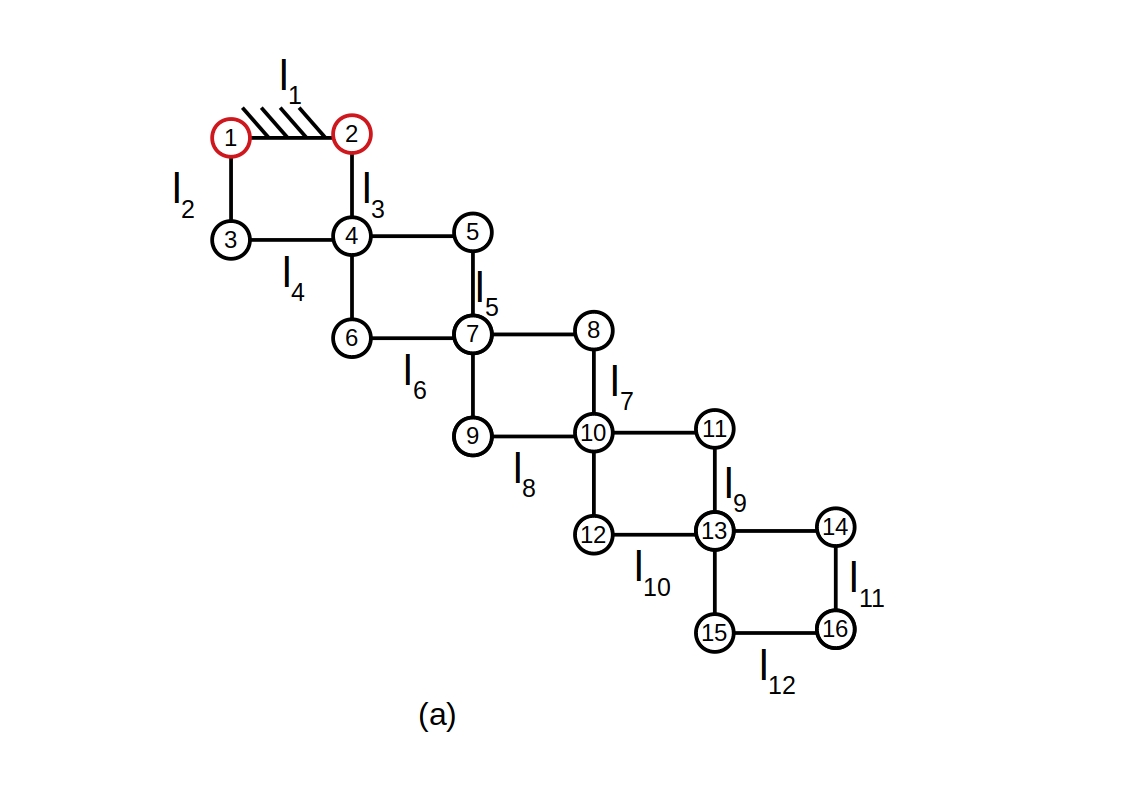} 
\includegraphics[scale=0.4]{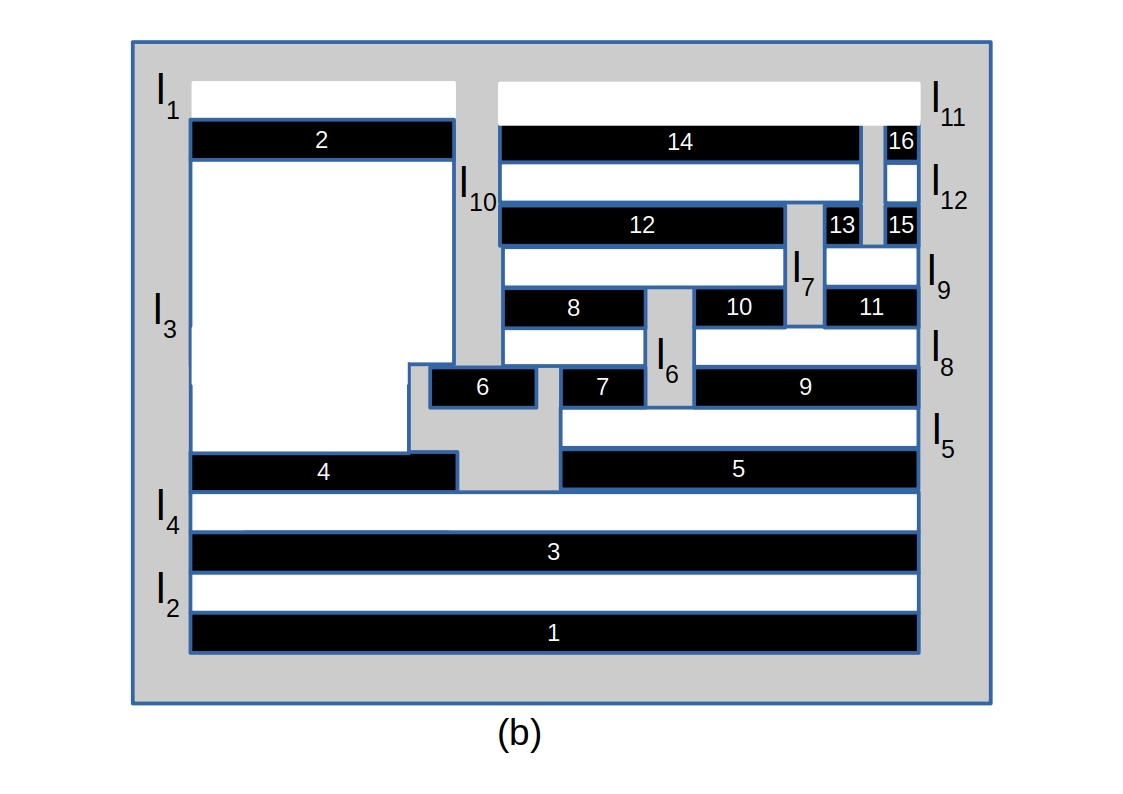} 
\caption{The zebra crossing algorithm applied to the staircase mechanism (a) The staircase mechanism (b) The zebra crossing diagram of the mechanism}
\label{fig_2}
\end{center}
\end{figure}
The staircase mechanism is shown in Figure \ref{fig_2} (a). The Zebra crossing diagram is drawn by following the \textit{Rule 1}, \textit{Rule 4}, \textit{Rule 6}. From here it forms two branches. For the first branch \textit{Rule 4}, followed by \textit{Rule 1} and \textit{Rule 4} is drawn. For the second branch, \textit{Rule 4} is applied. Then in the second branch the \textit{Rule 6} is applied twice. Then \textit{Rule 4} is applied to links $l_7$ and $l_9$. Then apply \textit{Rule 6} twice. To end the Zebra Crossing diagram, follow \textit{Rule 5}.


There are $16$ black patches and $12$ white patches. The number of loops (M) in the mechanism are calculated using the Equation \ref{equ_loop}, which is 
\begin{equation}
L=16-12+1=5
\end{equation}

The number of white patches between the black patches $N_w$ are $7$. The number of loops $L$ are $5$. The number of joints attached to the ground link $J_f$ are $2$. The degrees of freedom is calculated using the Equation \ref{equ_loopwhite},
\begin{equation}
M=7-5-2+1=1
\end{equation}

Therefore, the mechanism has $1$ degree of freedom.

\paragraph{Six-Bar Mechanism}
\begin{figure}[H]
\begin{center}
\includegraphics[scale=0.4]{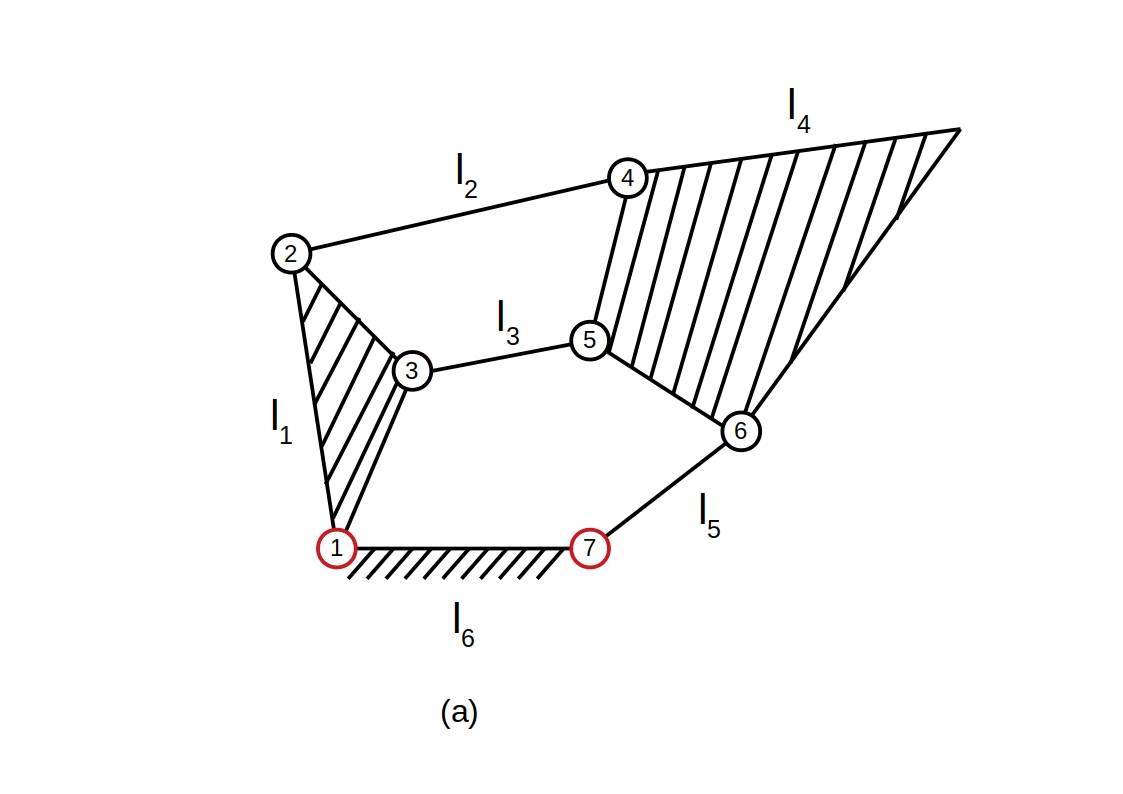} 
\includegraphics[scale=0.4]{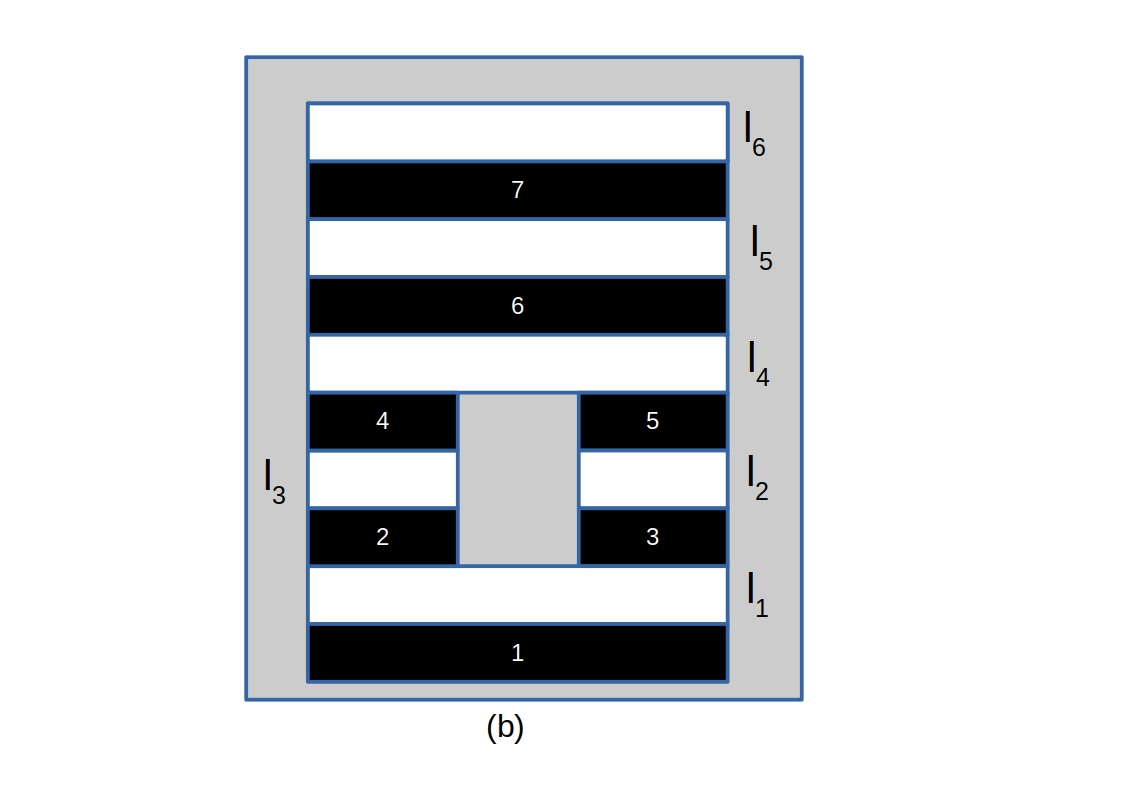} 
\caption{The zebra crossing algorithm applied to the six-bar mechanism (a) The six-bar mechanism (b) The zebra crossing diagram of the mechanism}
\label{fig_3}
\end{center}
\end{figure}
The six-bar mechanism is shown in Figure \ref{fig_3} (a). The mechanism has $6$ links and $7$ joints. The zebra crossing diagram for this mechanism is shown in Figure \ref{fig_3} (b). The drawing of the Zebra Crossing diagram starts by applying \textit{Rule 6}. Then on each branch apply \textit{Rule 4}. Then apply \textit{Rule 6}. This followed by \textit{Rule 4}, \textit{Rule 1} and \textit{Rule 4}.


There are $7$ black patches and $6$ white patches in the zebra crossing diagram. The number of loops in the mechanism are calculated using Equation \ref{equ_loop}. 
\begin{equation}
L=7-6+1=2
\end{equation}

The number of white patches between the black patches $N_w$ are $4$. The number of loops $L$ are $2$. The number of joints attached to the ground link $J_f$ are $2$. The degrees of freedom (M) is calculated using the Equation \ref{equ_loopwhite},
\begin{equation}
M=4-2-2+1=1
\end{equation}
Therefore, the mechanism has $1$ degree of freedom.

\paragraph{Five-Bar Mechanism}
\begin{figure}[H]
\begin{center}
\includegraphics[scale=0.4]{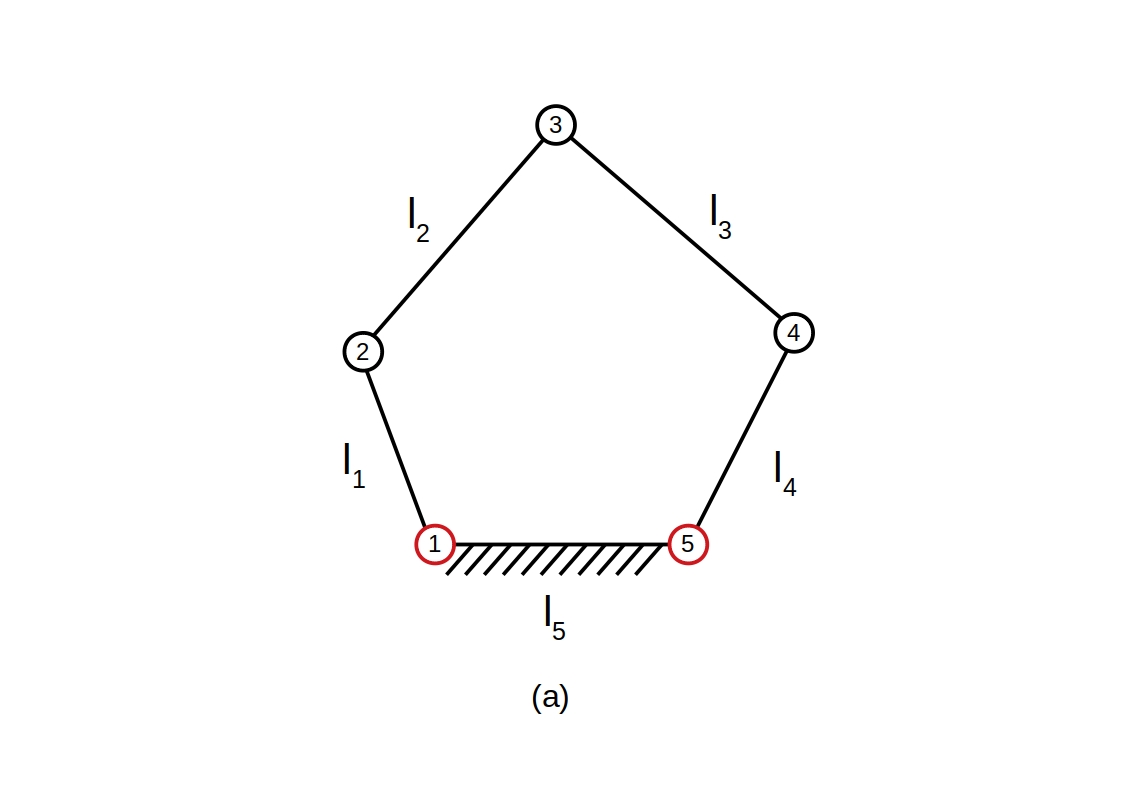} 
\includegraphics[scale=0.4]{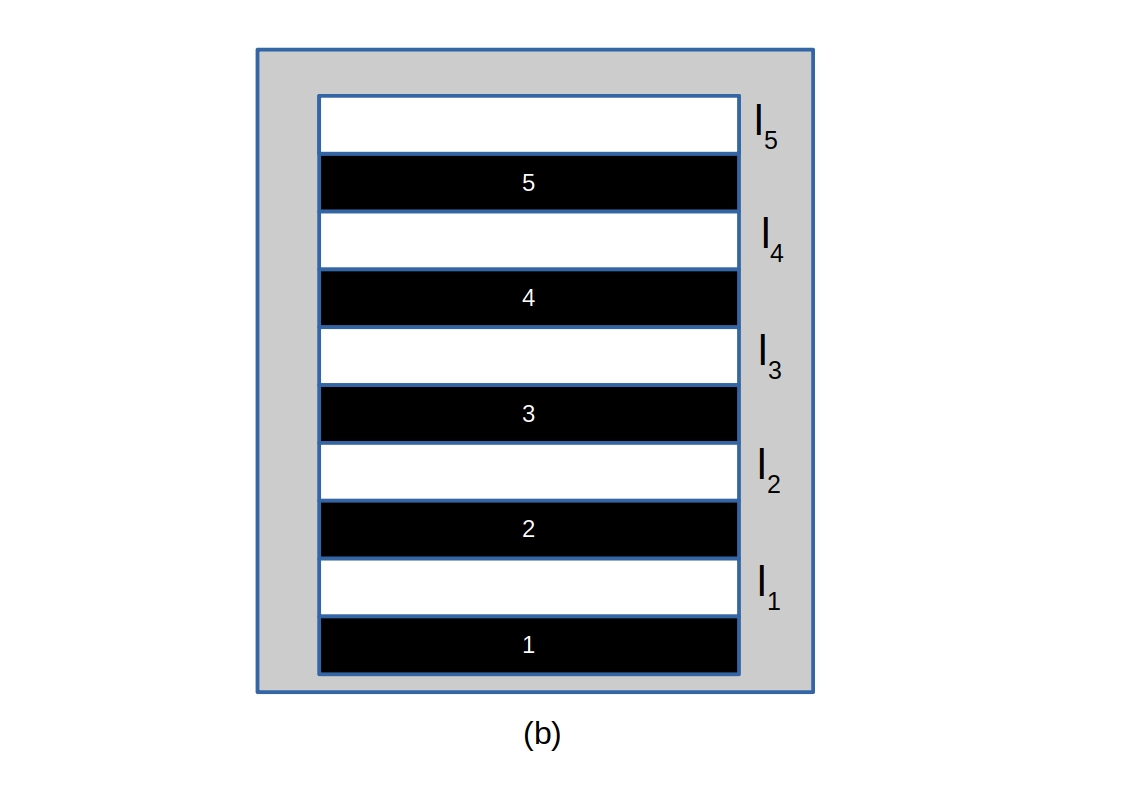} 
\caption{The zebra crossing algorithm applied to the five-bar mechanism (a) The five bar mechanism (b) The zebra crossing diagram of the mechanism}
\label{fig_4}
\end{center}
\end{figure}
The five-bar mechanism is shown in Figure \ref{fig_4} (a). The Zebra crossing diagram for this mechanism is shown in \ref{fig_4} (b). The steps involved in drawing the Zebra crossing diagram are similar to that of the four bar mechanism. In this case an additional link and joint are added. 

There are $5$ black patches and $5$ white patches in the zebra crossing diagram. The number of loops in the mechanism are calculated using the Equation \ref{equ_loop}. 
\begin{equation}
L=5-5+1=1
\end{equation}

The number of white patches between the black patches $N_w$ are $4$. The number of loops $L$ is $1$. The number of joints attached to the ground link $J_f$ are $2$. The degrees of freedom (M) is calculated using the Equation \ref{equ_loopwhite},
\begin{equation}
M=4-1-2+1=2
\end{equation}
Therefore, the mechanism has $2$ degrees of freedom.

\paragraph{Four-Bar Mechanism in Series}
\begin{figure}[H]
\begin{center}
\includegraphics[scale=0.4]{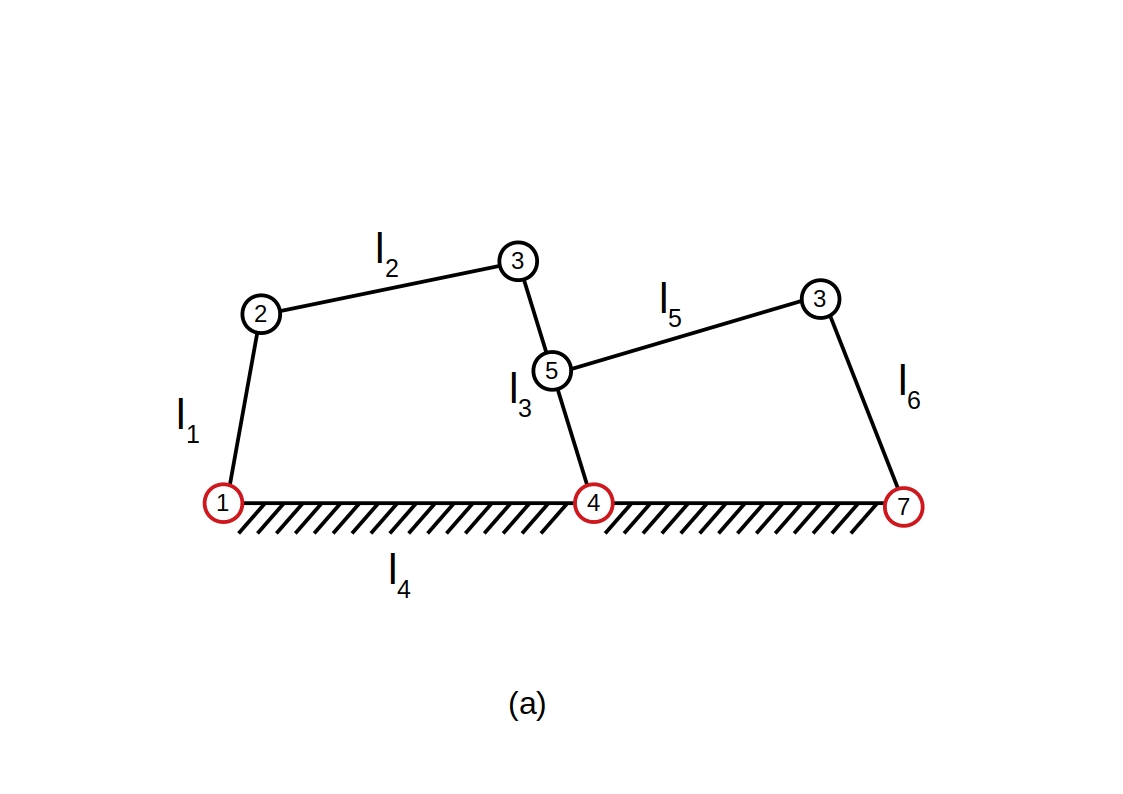} 
\includegraphics[scale=0.4]{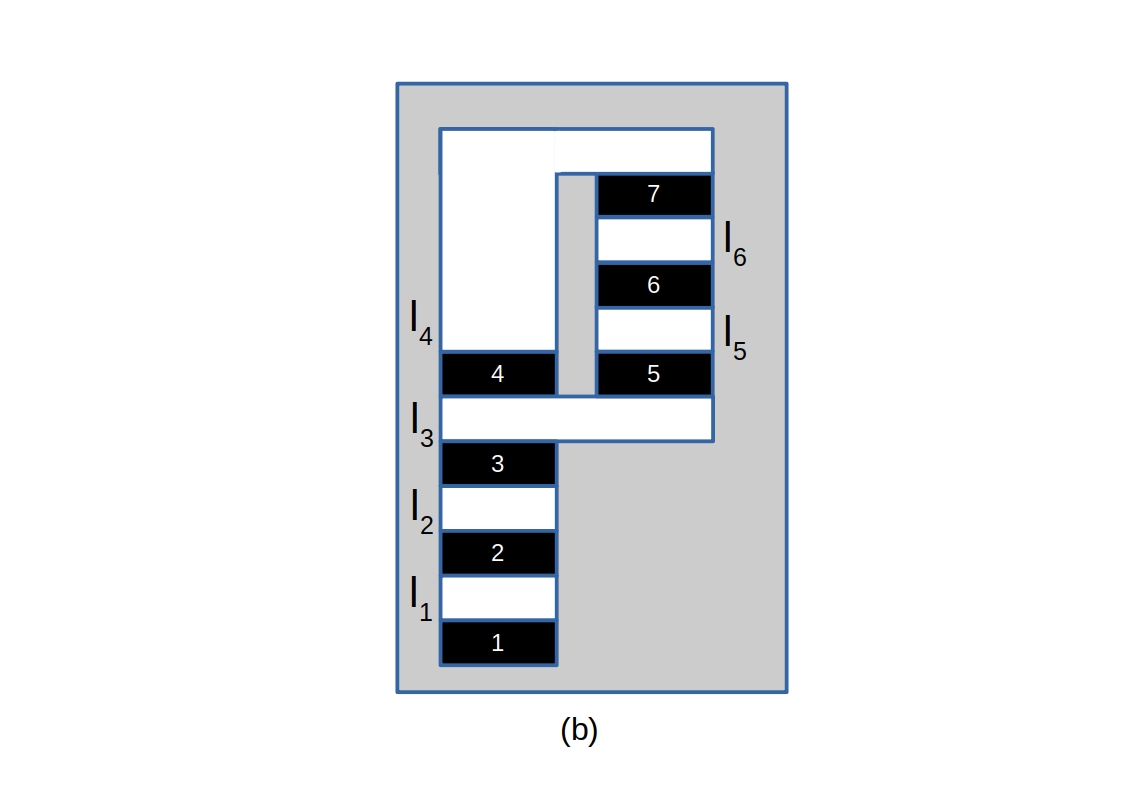} 
\caption{The zebra crossing algorithm applied to the four-bar mechanism in series (a) The four-bar mechanism in series (b) The zebra crossing diagram of the mechanism}
\label{fig_5}
\end{center}
\end{figure}
The four-bar mechanism in series is shown in Figure \ref{fig_5} (a). The Zebra crossing diagram for this mechanism is shown in Figure \ref{fig_5} (b). The steps followed are similar to that of a four bar mechanism but the link $l_3$ is ternary in nature and hence \textit{Rule 6} has to be applied. 

There are $7$ black patches and $6$ white patches in the zebra crossing diagram. The number of loops (L) in the mechanism are calculated using the Equation \ref{equ_loop}. 
\begin{equation}
L=7-6+1=2
\end{equation}

The number of white patches between the black patches $N_w$ are $5$. The number of loops $L$ are $2$. The number of joints attached to the ground link $J_f$ are $3$. The degrees of freedom is calculated using the Equation \ref{equ_loopwhite},
\begin{equation}
M=5-2-3+1=1
\end{equation}
Therefore the mechanism has $1$ degree of freedom.

\paragraph{Crank and Slider Mechanism}
\begin{figure}[H]
\begin{center}
\includegraphics[scale=0.4]{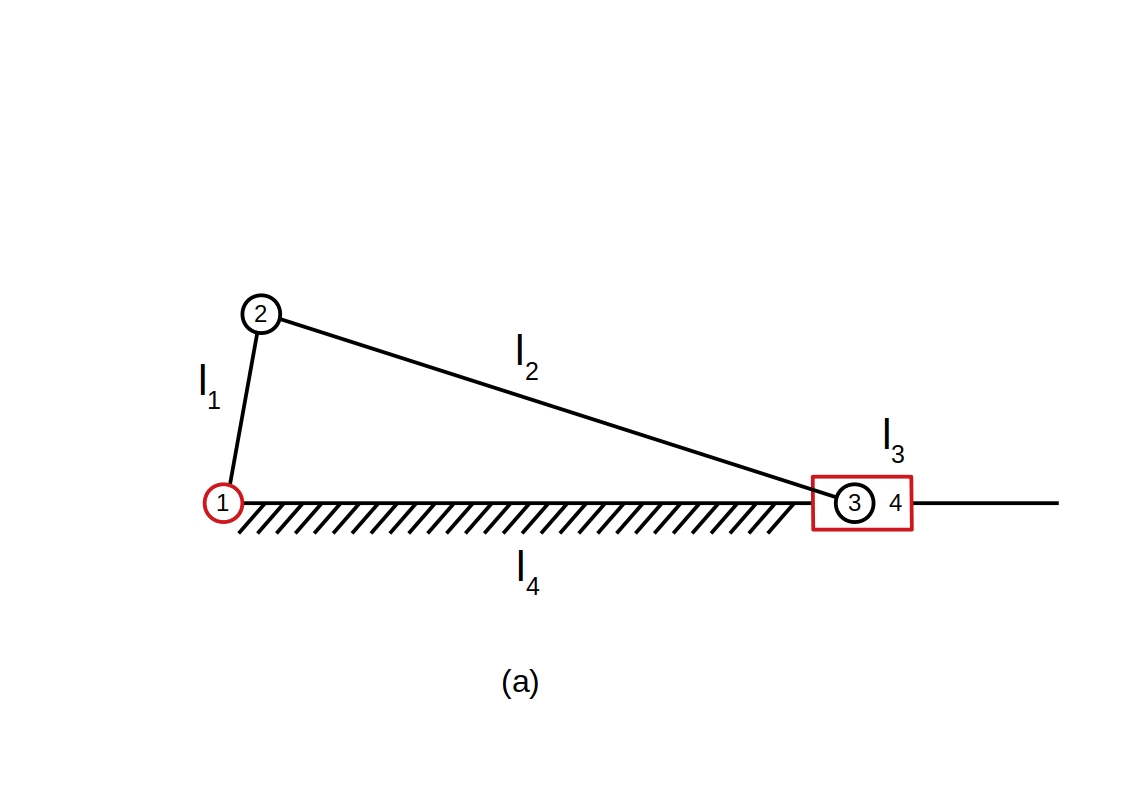} 
\includegraphics[scale=0.4]{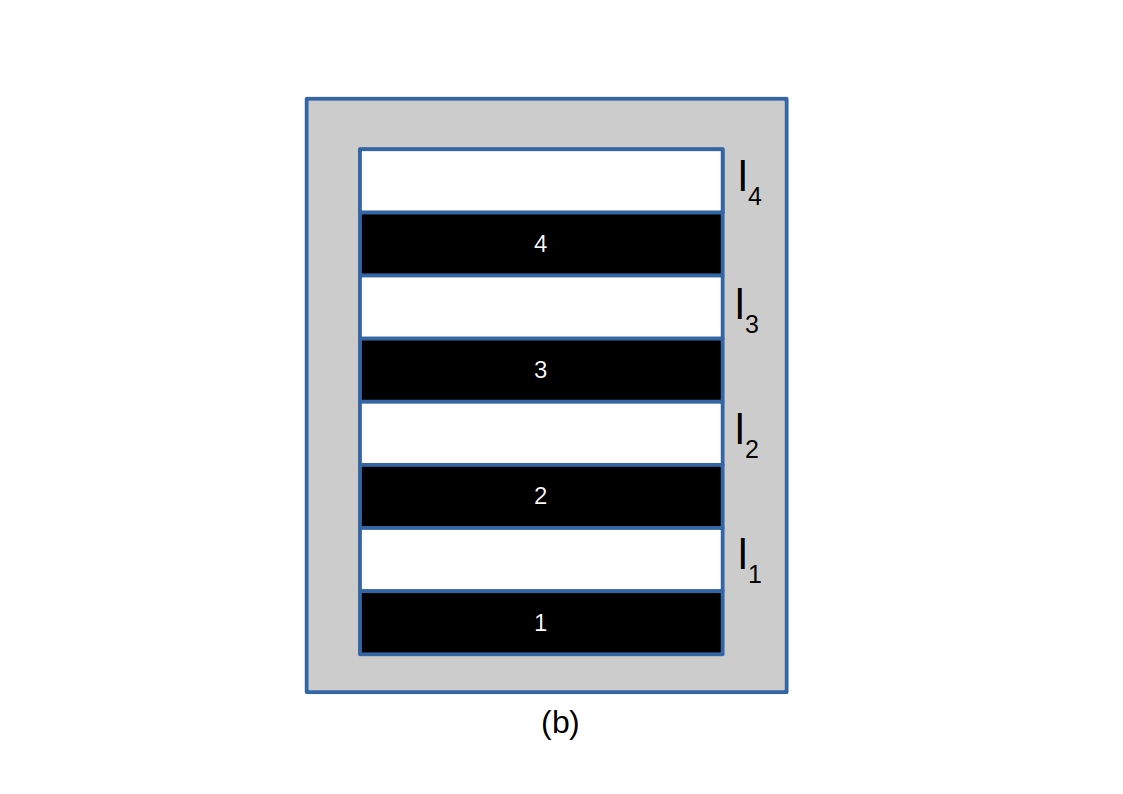} 
\caption{The zebra crossing algorithm applied to the crank and slider mechanism (a) The crank and slider mechanism (b) The zebra crossing diagram of the mechanism}
\label{fig_6}
\end{center}
\end{figure}
The crank and slider mechanism is shown in Figure \ref{fig_6} (a). This mechanism has $4$ links and $4$ joints. The number of revolute joints are $3$ and prismatic joint is $1$. The zebra crossing diagram for this mechanism is shown in Figure \ref{fig_6} (b). The steps involved in drawing the zebra crossing diagram is similar to that of the four bar mechanism.  

There are $4$ black patches and $4$ white patches in the zebra crossing diagram. The number of loops in the mechanism are calculated using Equation \ref{equ_loop}. The number of loops are,
\begin{equation}
L=4-4+1=1
\end{equation}

The number of white patches between the black patches $N_w$ are $3$. The number of loops $L$ is $1$. The number of joints attached to the ground link $J_f$ are $2$. The degrees of freedom is calculated using Equation \ref{equ_loopwhite},
\begin{equation}
M=3-1-2+1=1
\end{equation}
Therefore, the mechanism has $1$ degree of freedom.

\paragraph{Four-bar Mechanism in Parallel}
\label{example_four_bar}
\begin{figure}[H]
\begin{center}
\includegraphics[scale=0.4]{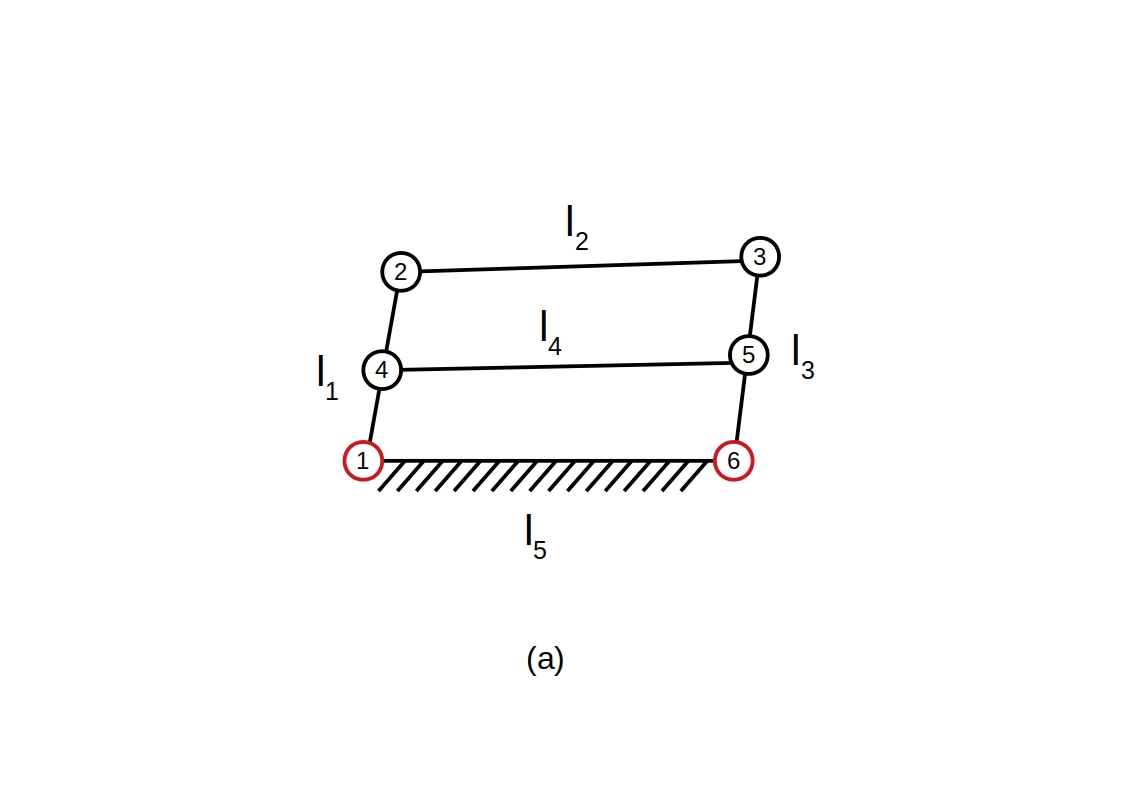} 
\includegraphics[scale=0.4]{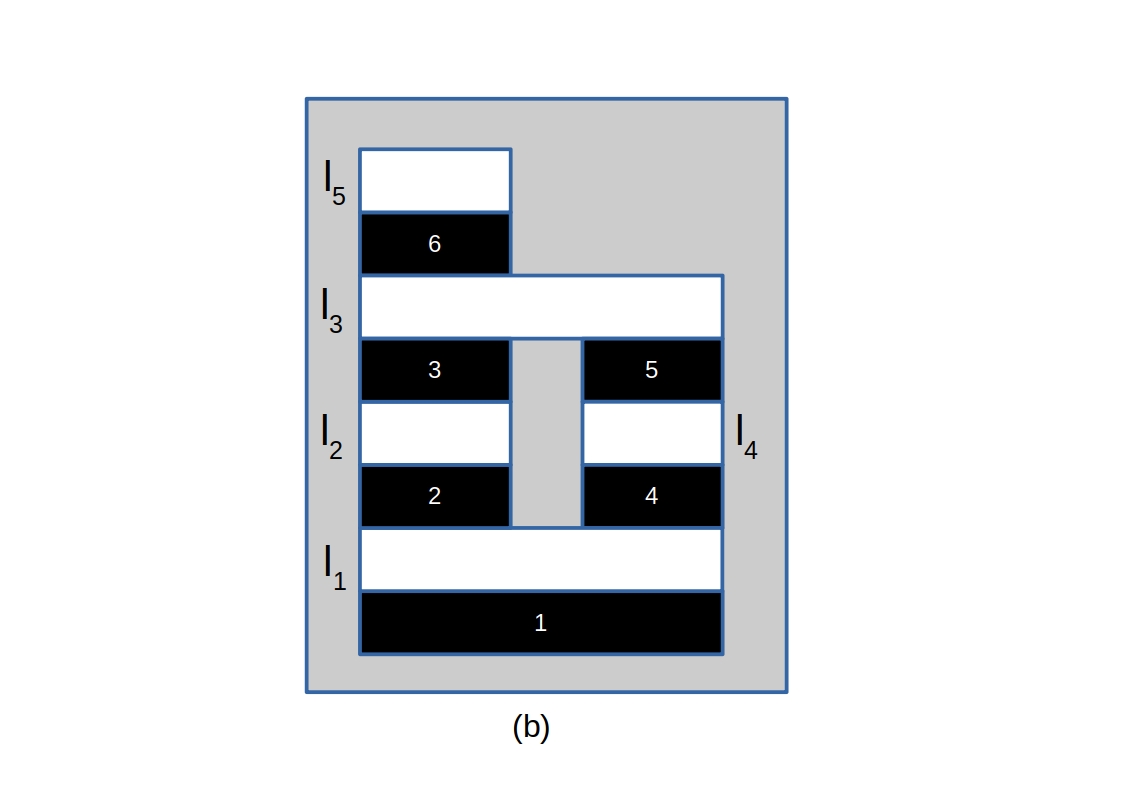} 
\caption{The zebra crossing algorithm applied to the four-bar mechanism in parallel (a) The four-bar mechanism in parallel (b) The zebra crossing diagram of the mechanism}
\label{fig_7}
\end{center}
\end{figure}

The four-bar mechanism in parallel is shown in Figure \ref{fig_7} (a). The Zebra crossing diagram is shown in Figure \ref{fig_7} (b). The steps involved in drawing the Zebra crossing diagram are similar to that of the six bar mechanism. The degree of freedom for this mechanism depends on the length of the links $l_2$, $l_4$ and $l_5$. There are two cases which arises based on the length of the mechanism. 

\textit{Case 1}: The links $l_2$, $l_4$ and $l_5$ are of same length. 

There are $6$ black patches and $5$ white patches in the zebra crossing diagram. The number of loops in the mechanism are calculated using Equation \ref{equ_loop}. The number of loops are,
\begin{equation}
L=6-5+1=2
\end{equation}

The number of white patches between the black patches $N_w$ are $4$. The number of loops $L$ is $2$. The number of joints attached to the ground link $J_f$ are $2$. The degrees of freedom is calculated using Equation \ref{equ_loopwhite}.
Calculating the degrees of freedom,
\begin{equation}
M=4-2-2+1=1
\end{equation}
Therefore, the mechanism has $1$ degree of freedom. 

\textit{Case 2}: The links $l_2$, $l_4$ and $l_5$ are of different lengths. 

There are $6$ black patches and $5$ white patches in the zebra crossing diagram. The number of loops (L) in the mechanism are calculated using Equation \ref{equ_loop}.
\begin{equation}
L=6-5+1=2
\end{equation}

The number of white patches between the black patches $N_w$ are $3$ (the links $l_2$ and $l_4$ are treated as one white patch). The number of loops $L$ are $2$. The number of joints attached to the ground link $J_f$ are $2$. The degrees of freedom (M) is calculated using Equation \ref{equ_loopwhite}.

\begin{equation}
M=3-2-2+1=0
\end{equation}
Therefore the mechanism has $0$ degree of freedom. 

The degrees of freedom of a mechanism will vary based on the link lengths and joint ranges. This variation will be taken in to account and the degrees of freedom will be calculated in our future works.

\subsubsection{Spatial mechanisms}
\paragraph{Stewart Platform}
\begin{figure}[H]
\begin{center}
\includegraphics[scale=0.4]{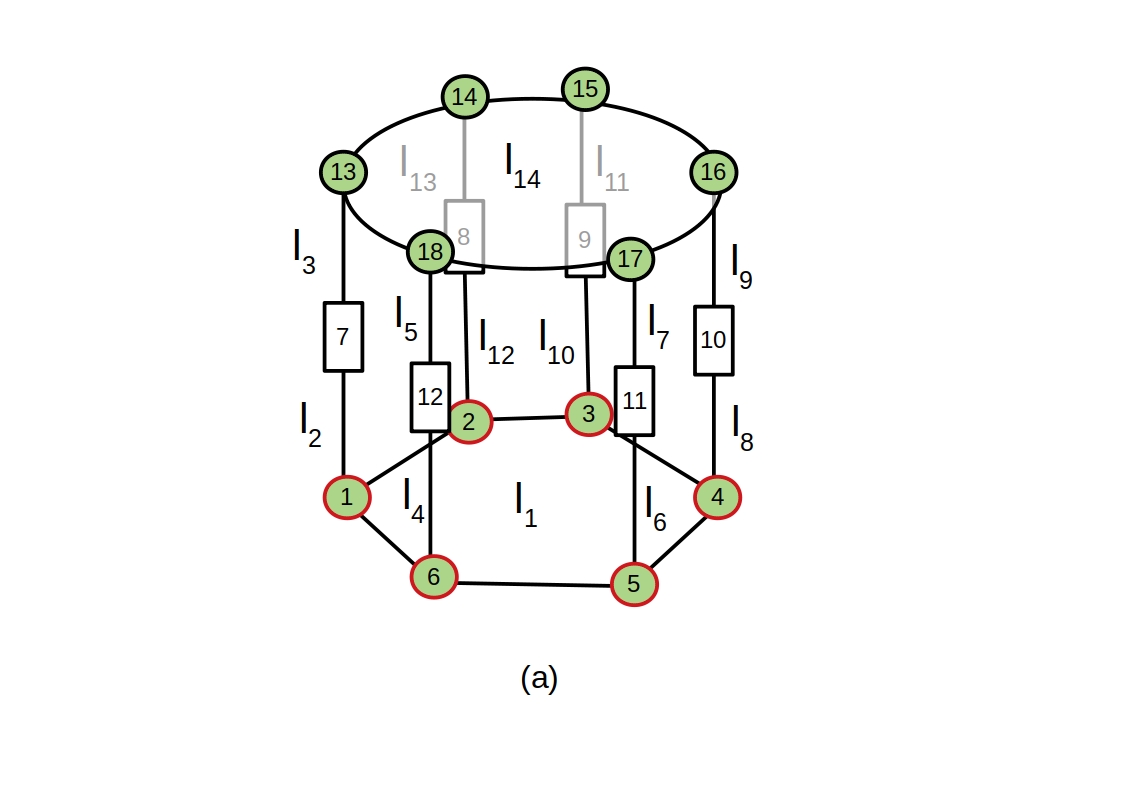} 
\includegraphics[scale=0.4]{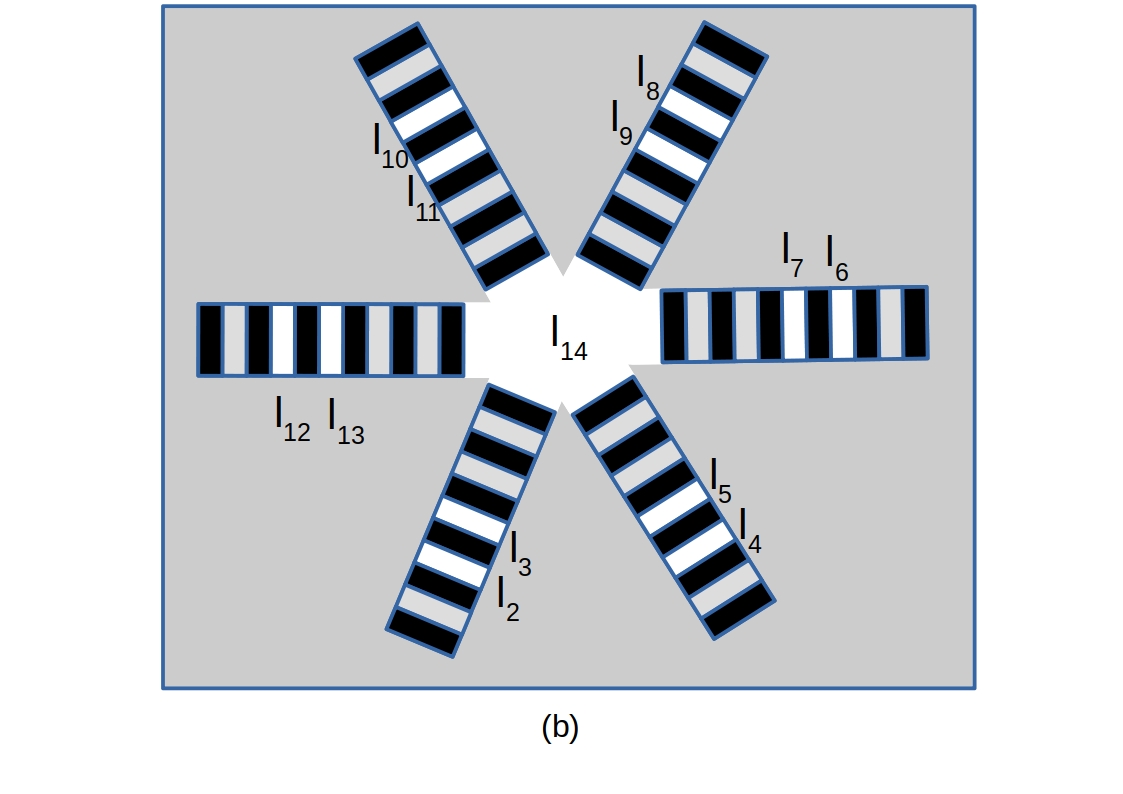}  
\caption{The zebra crossing algorithm applied to the Stewart platform (a) The Stewart platform (b) The zebra crossing diagram of the mechanism}
\label{fig_9}
\end{center}
\end{figure}
The Stewart platform is shown in Figure \ref{fig_9} (a). The link $l_1$ is the fixed link and the link $l_{14}$ is the moving platform. There are $6$ legs which connect the fixed link to the moving platform. To draw the Zebra crossing diagram for this mechanism, one of the legs is considered. The \textit{Rule 2} is applied to start drawing the Zebra Crossing diagram, which is followed by \textit{Rule 4} and \textit{Rule 3}. This is one leg of the parallel manipulator. This terminates at the movable platform which is the hexanary link. Similarly, the other \textit{5} legs of the parallel manipulator are drawn. 

There are $36$ black patches and $31$ white and grey patches in the zebra crossing diagram. The number of loops (L) in the mechanism are found out using the Equation \ref{equ_loop}.
\begin{equation}
L=36-31+1=6
\end{equation}

The number of white patches between the black patches $N_w$ are $17$. It is calculated as follows: The number of white patches between the black patches in the legs are $12$. The link $l_{14}$, which is the white patch, is assigned a number that is one less than the number of legs in the parallel manipulator, which is $5$. This is a rule that has to be followed for every parallel manipulator. The number of loops $L$ are $6$. The number of joints attached to the ground link $J_f$ are $6$. The degrees of freedom (M) is calculated using the Equation \ref{equ_loopwhite}.
\begin{equation}
M=17-6-6+1=6
\end{equation}
Therefore the mechanism has $6$ degrees of freedom.

\paragraph{Cartesian Parallel Manipulator}
\begin{figure}[H]
\begin{center}
\includegraphics[scale=0.4]{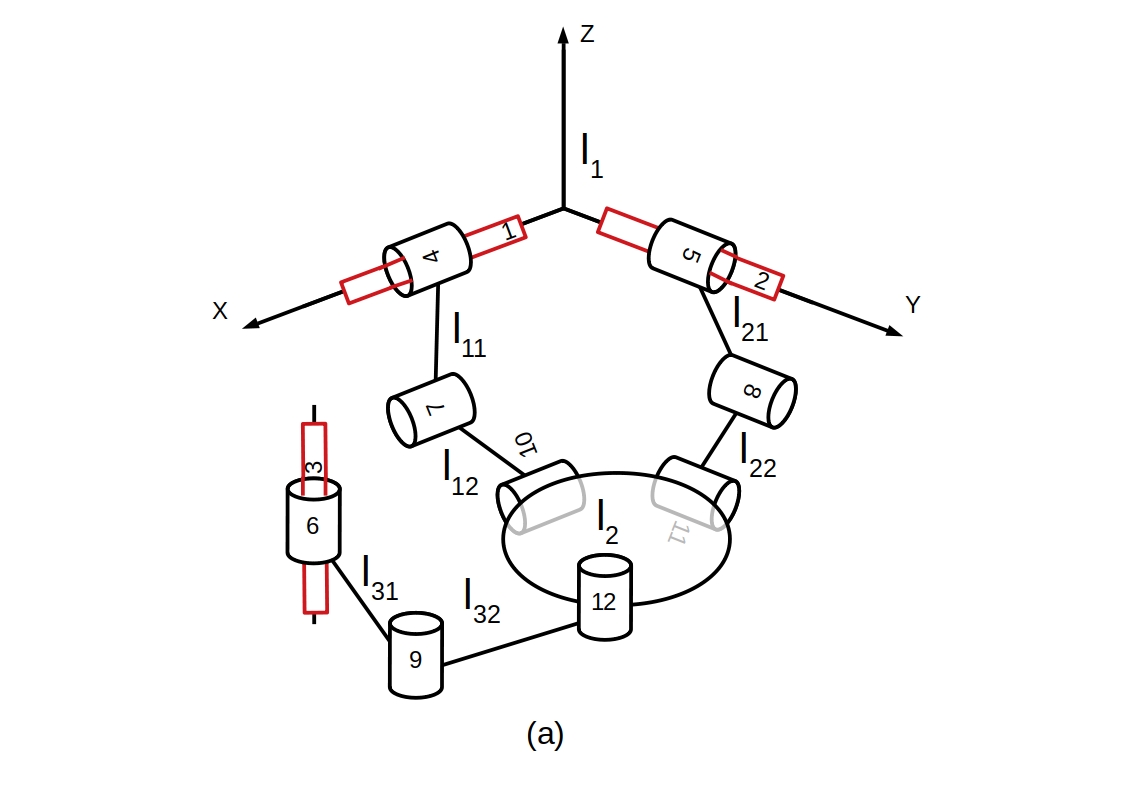} 
\includegraphics[scale=0.4]{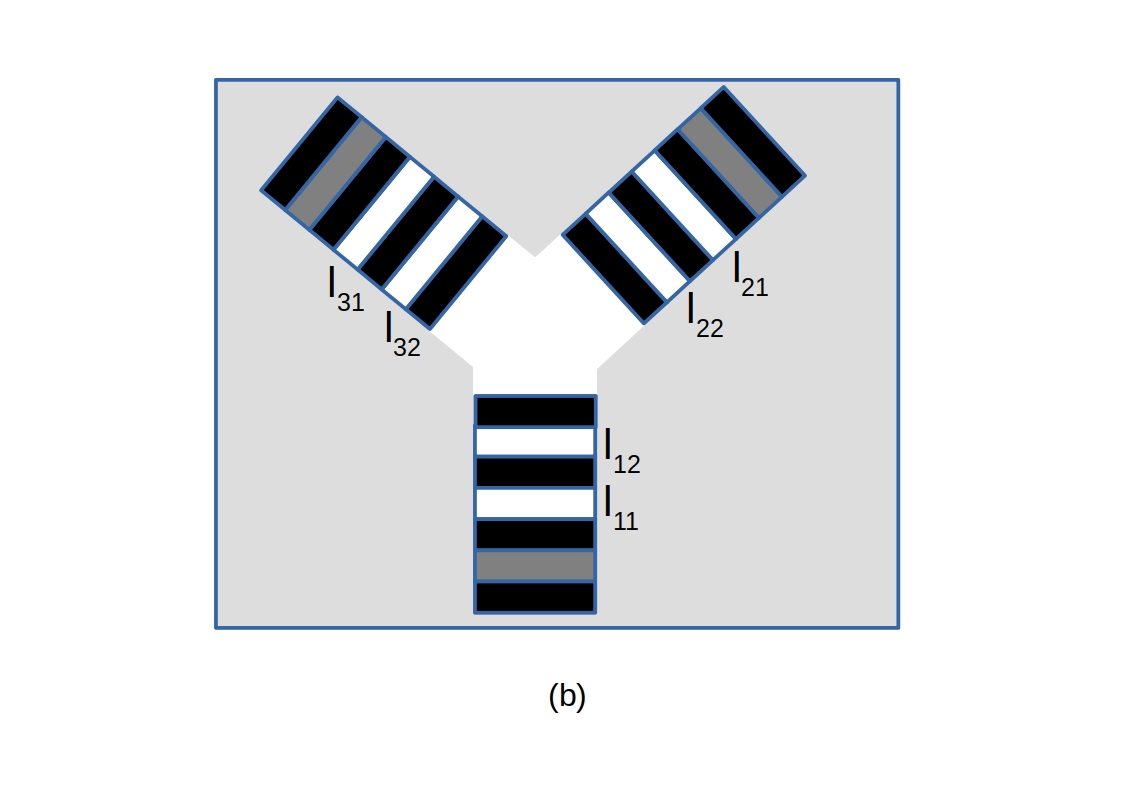}  
\caption{The zebra crossing algorithm applied to the cartesian parallel manipulator (a) The cartesian parallel manipulator (b) The zebra crossing diagram of the mechanism}
\label{fig_10}
\end{center}
\end{figure}
The cartesian parallel manipulator is shown in Figure \ref{fig_10} (a). The link $l_1$ is the fixed link and the link $l_2$ is the moving platform. They are connected by three legs. The Zebra crossing diagram is shown in Figure \ref{fig_10} (b). The Zebra Crossing is drawn by applying \textit{Rule 2} followed by \textit{Rule 4}, \textit{Rule 1}, \textit{Rule 4} and \textit{Rule 1}. This is one leg of the manipulator. This leg terminates at the ternary link $l_2$. Similarly, the other \textit{2} legs of the parallel manipulator are drawn.


There are $12$ black patches and $10$ white and grey patches in the zebra crossing diagram. The number of loops (L) in the mechanism are calculated using the Equation \ref{equ_loop}.
\begin{equation}
L=12-10+1=3
\end{equation}

The number of white patches between the black patches $N_w$ are $8$ (the number of white patches between the black patches in the legs are $6$ and for the link $l_{2}$, the white patch is assigned a number which is one less than the number of legs in the parallel manipulator, which is $2$). The number of loops $L$ are $3$. The number of joints attached to the ground link $J_f$ are $3$. The degrees of freedom is calculated using the Equation \ref{equ_loopwhite}.
\begin{equation}
M=8-3-3+1=3
\end{equation}
Therefore the mechanism has $3$ degrees of freedom.

\paragraph{Orthoglide Manipulator}
\begin{figure}[H]
\begin{center}
\includegraphics[scale=0.4]{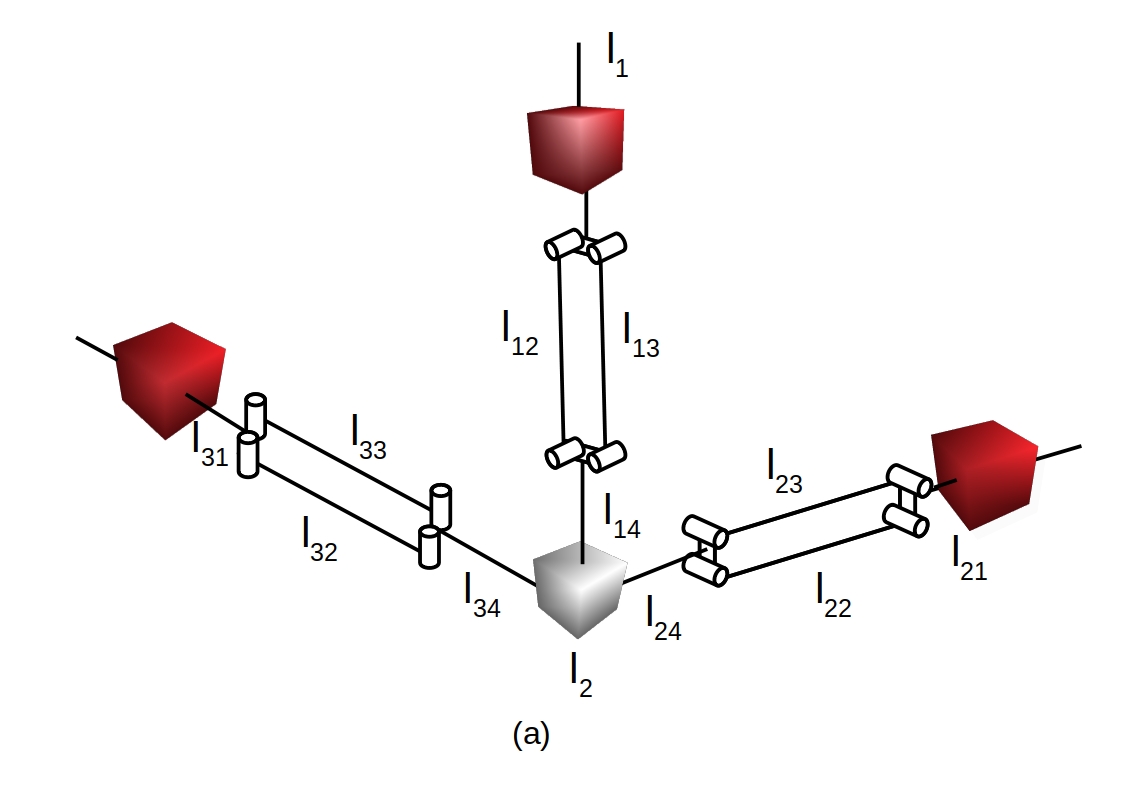} 
\includegraphics[scale=0.4]{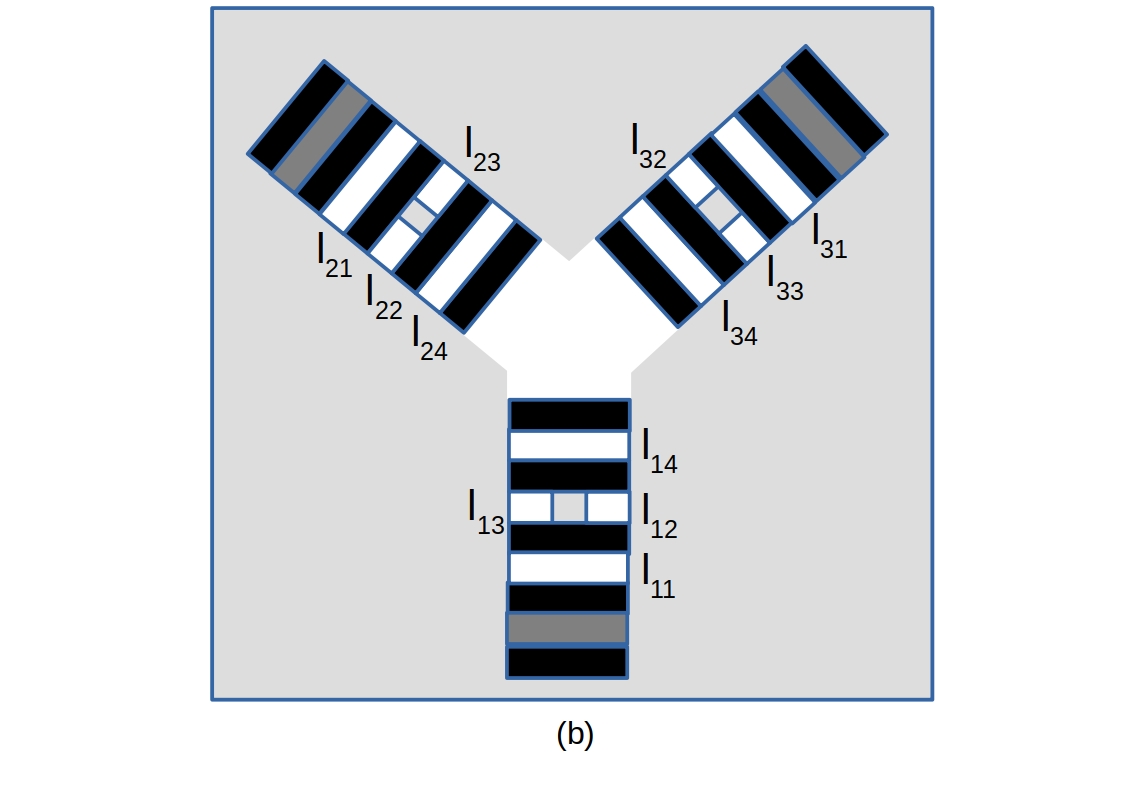} 
\caption{The zebra crossing algorithm applied to the orthoglide manipulator (a) The orthoglide manipulator (b) The zebra crossing diagram of the mechanism}
\label{fig_11}
\end{center}
\end{figure}
The orthoglide manipulator is shown in Figure \ref{fig_11} (a). The link $l_1$ is the fixed link and the link $l_2$ is the moving platform. There are three legs which connect the fixed link and the moving platform. The Zebra crossing diagram is drawn as follows. The \textit{Rule 2} is applied to start with, followed by \textit{Rule 4} and \textit{Rule 1}. The \textit{Rule 4} is applied twice to form two white patches. This is followed by \textit{Rule 5}, which connects to the movable platform. Similarly, the other \textit{2} legs of the parallel manipulator are drawn.


There are $21$ black patches and $16$ white and grey patches in the zebra crossing diagram. The number of loops (L) in the mechanism are calculated using the Equation \ref{equ_loop}. 
\begin{equation}
L=21-16+1=6
\end{equation}

The number of white patches between the black patches $N_w$ are $11$ (the number of white patches between the black patches in the legs are $9$ and for the link $l_{2}$, the white patch is assigned a number which is one less than the number of legs in the parallel manipulator, which is $2$). The number of loops $L$ are $6$. The number of joints attached to the ground link $J_f$ are $3$. The degrees of freedom (M) is calculated using Equation \ref{equ_loopwhite}.

\begin{equation}
M=11-6-3+1=3
\end{equation}
Therefore the mechanism has $3$ degrees of freedom.

\paragraph{H4 Parallel Manipulator}
\begin{figure}[H]
\begin{center}
\includegraphics[scale=0.4]{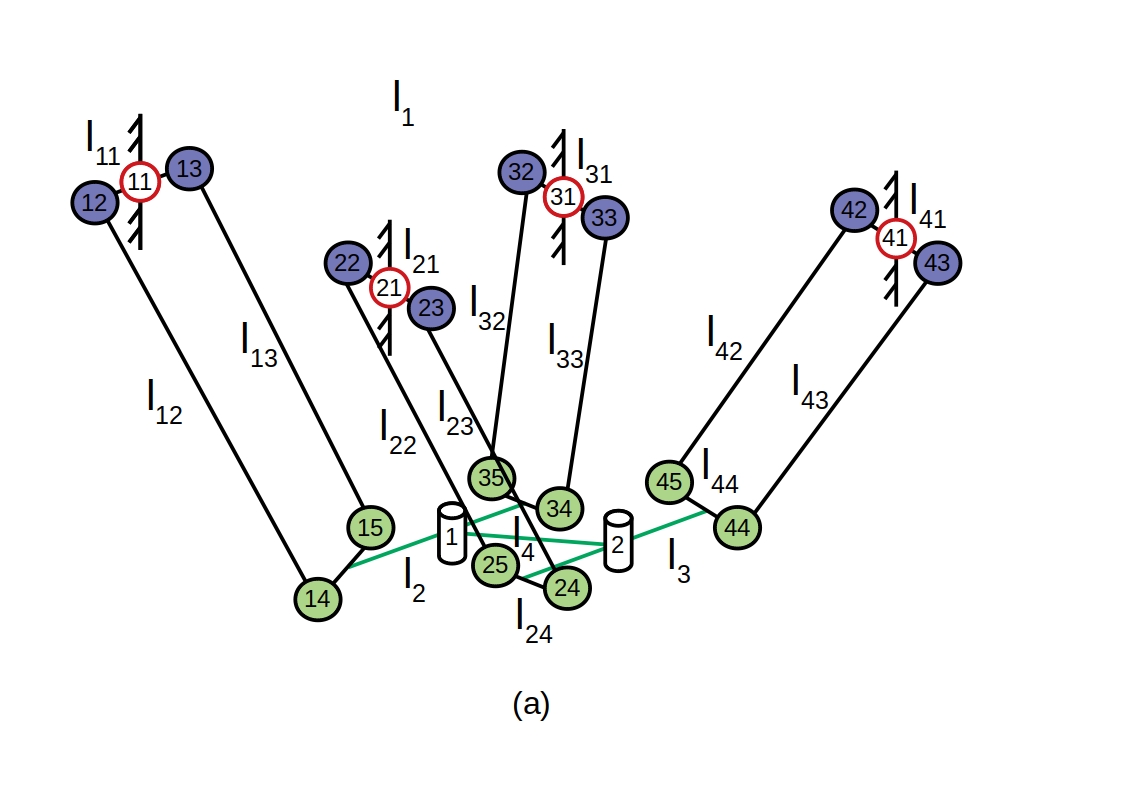} 
\includegraphics[scale=0.4]{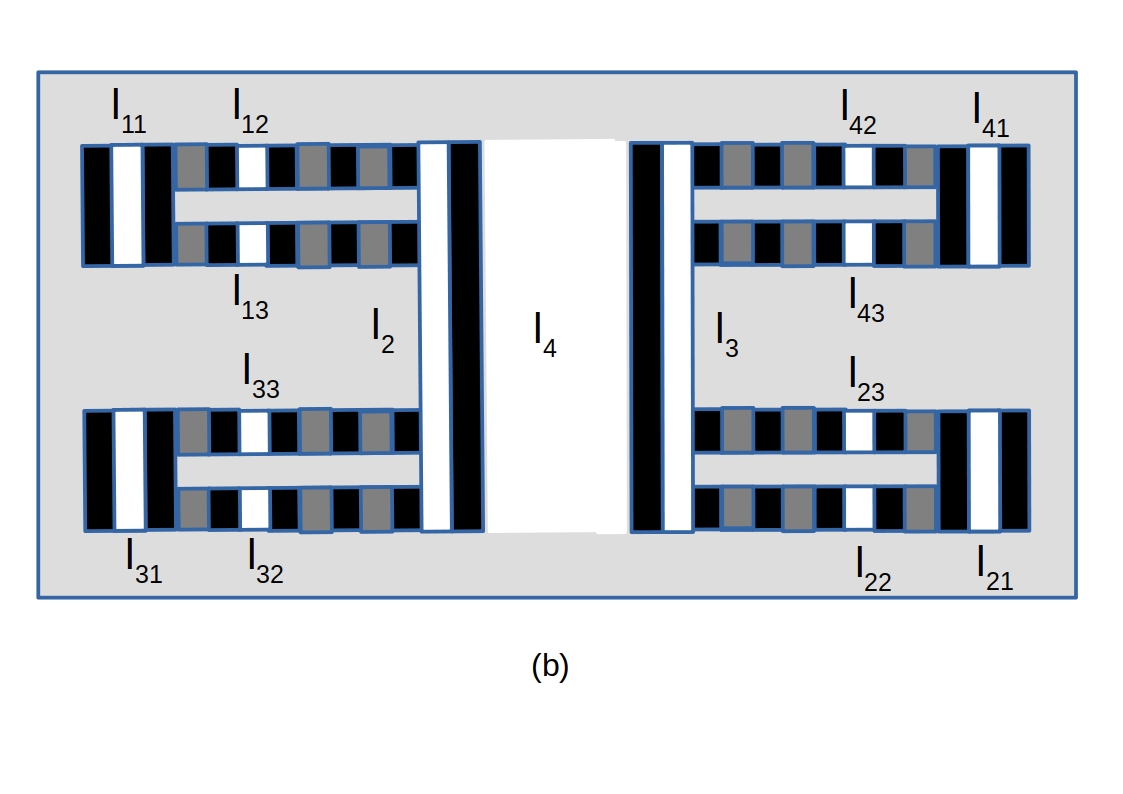}  
\caption{The zebra crossing algorithm applied to the H4 parallel manipulator (a) The H4 parallel manipulator (b) The zebra crossing diagram of the mechanism}
\label{fig_12}
\end{center}
\end{figure}
The H4 Parallel Manipulator is shown in Figure \ref{fig_12} (a). This manipulator has a fixed link $l_1$ and a moving platform $l_4$. It has four legs which connect the fixed link and the moving platform. The Zebra crossing diagram is drawn as follows. The \textit{Rule 1} is applied, which is followed by \textit{Rule 4}. Now the branching happens where the \textit{Rule 2}, \textit{Rule 4} and \textit{Rule 3} are applied in two branches. Similarly, the second is drawn. The two legs are connected using \textit{Rule 4} and \textit{Rule 1} to the movable platform $l_4$. Similarly, the other \textit{2} legs of the parallel manipulator are drawn.


There are $46$ black patches and $39$ white and grey patches in the zebra crossing diagram. The number of loops in the mechanism are calculated using Equation \ref{equ_loop}. 
\begin{equation}
L=46-39+1=8
\end{equation}

The number of white patches between the black patches $N_w$ are $15$. The number of loops $L$ are $8$. The number of joints attached to the ground link $J_f$ are $4$. The degrees of freedom is calculated using Equation \ref{equ_loopwhite}.
\begin{equation}
M=15-8-4+1=4
\end{equation}
Therefore, the mechanism has $4$ degrees of freedom.

\paragraph{STAR Manipulator}
\begin{figure}[H]
\begin{center}
\includegraphics[scale=0.4]{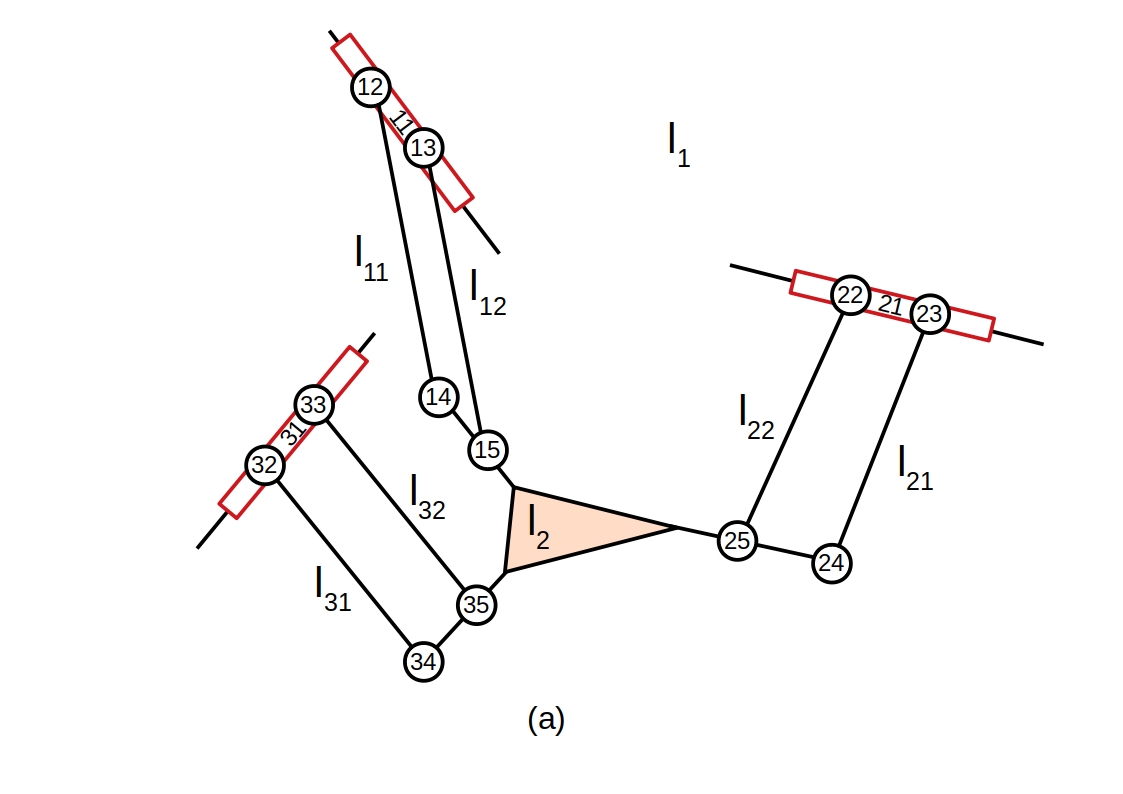} 
\includegraphics[scale=0.4]{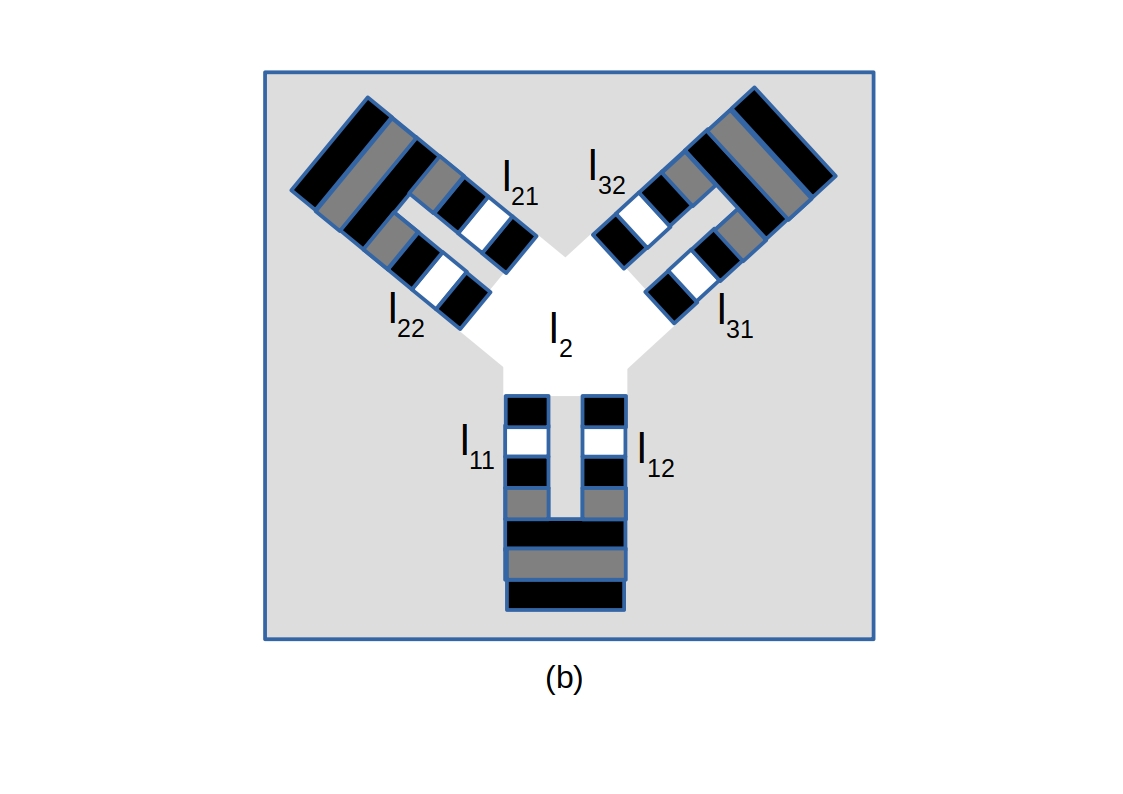} 
\caption{The zebra crossing algorithm applied to the STAR manipulator (a) The STAR manipulator (b) The zebra crossing diagram of the mechanism}
\label{fig_13}
\end{center}
\end{figure}
The STAR manipulator is shown in Figure \ref{fig_13} (a). The link $l_{1}$ is the fixed link and the link $l_{2}$ is the moving platform. The link $l_{1}$ is connected to the link $l_{2}$ through three legs. The Zebra crossing diagram for this manipulator is shown in Figure \ref{fig_13} (b). One leg of the manipulator is drawn as follows. Apply \textit{Rule 2} which represents the cylindrical joint. This joint branches in to two and the \textit{Rule 2} is applied. Then the \textit{Rule 4}  followed by \textit{Rule 1} and \textit{Rule 9}. Similarly, the other two legs are represented in Zebra crossing diagram.

 
There are $21$ black patches and $16$ white and grey patches in the zebra crossing diagram. The number of loops in the mechanism are calculated using the Equation \ref{equ_loop}. 
\begin{equation}
L=21-16+1=6
\end{equation}

The number of white patches between the black patches $N_w$ are $11$ (the number of white patches between the black patches in the legs are $9$ and for the link $l_{2}$, the white patch is assigned a number which is one less than the number of legs in the parallel manipulator, which is $2$). The number of loops $L$ are $6$. The number of joints attached to the ground link $J_f$ are $3$. The degrees of freedom (M) is calculated using the Equation \ref{equ_loopwhite}.
\begin{equation}
M=11-6-3+1=3
\end{equation}
Therefore the mechanism has $3$ degrees of freedom.

\paragraph{DELTA Manipulator}
\begin{figure}[H]
\begin{center}
\includegraphics[scale=0.4]{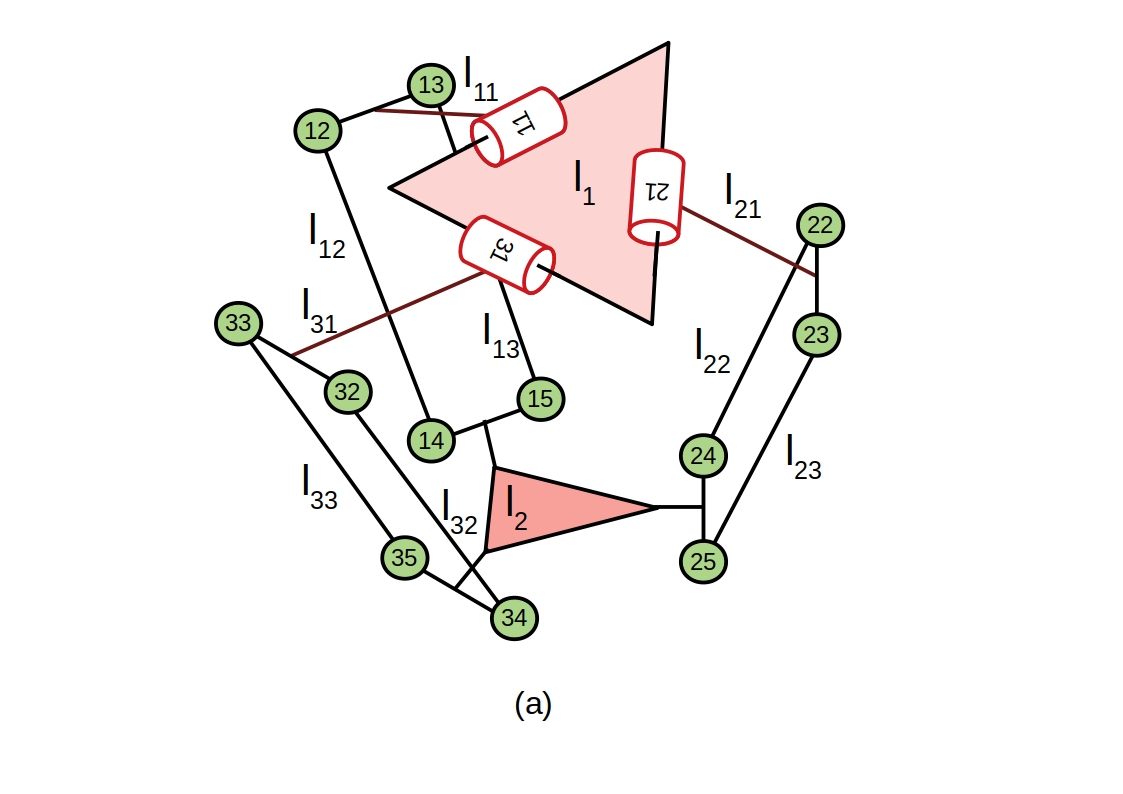} 
\includegraphics[scale=0.4]{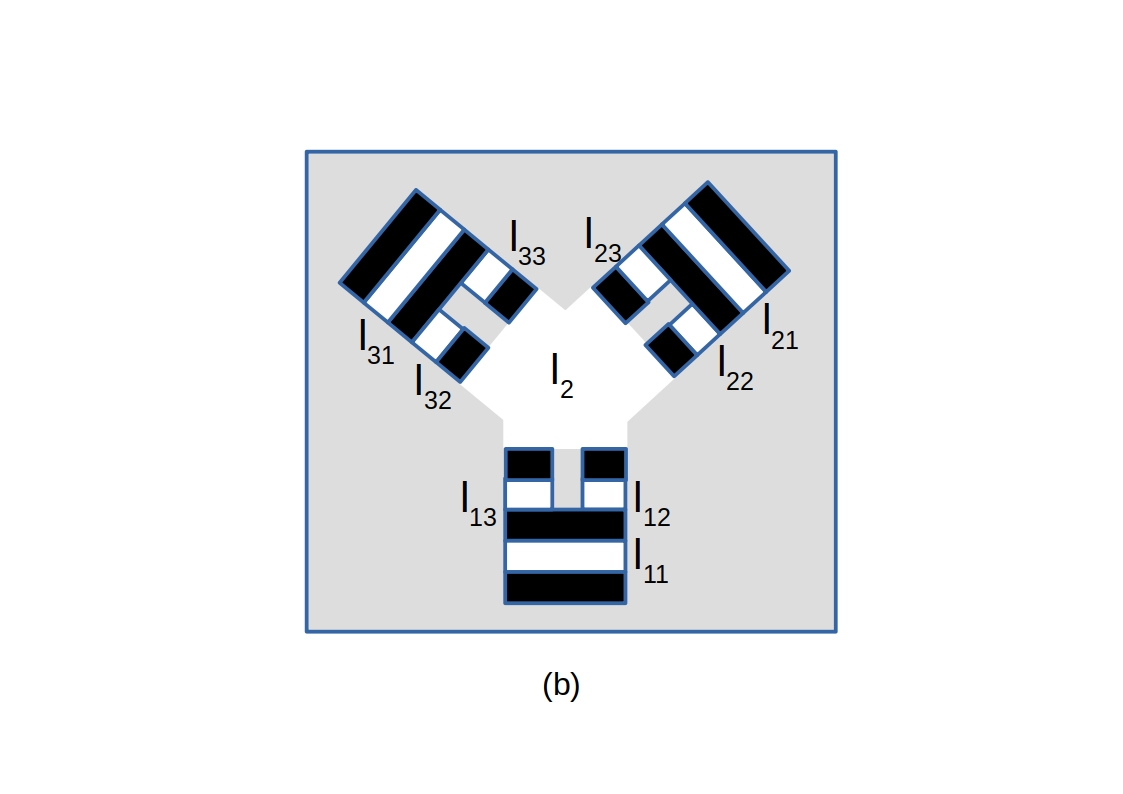} 
\caption{The zebra crossing algorithm applied to the DELTA manipulator (a) The DELTA manipulator (b) The zebra crossing diagram of the mechanism}
\label{fig_14}
\end{center}
\end{figure}
The DELTA manipulator is shown in Figure \ref{fig_14} (a). The link $l_1$ is fixed and the link $l_2$ is the moving platform. There are three legs which connect the fixed link to the moving platform. The Zebra crossing diagram for this mechanism is shown in Figure \ref{fig_14} (b). It is drawn by applying \textit{Rule 1} followed by \textit{Rule 4}. From this, two links branch out which is drawn by applying \textit{Rule 5} and ends by applying \textit{Rule 9}. Similarly, the other two legs are drawn.


There are $15$ black patches and $10$ white patches in the zebra crossing diagram. The number of loops (L) in the mechanism are calculated using the Equation \ref{equ_loop}. 
\begin{equation}
L=15-10+1=6
\end{equation}

The number of white patches between the black patches $N_w$ are $11$ (the number of white patches between the black patches in the legs are $6$ and for the link $l_{2}$, the white patch is assigned a number which is one less than the number of legs in the parallel manipulator, which is $5$). The number of loops $L$ are $6$. The number of joints attached to the ground link $J_f$ are $3$. The degrees of freedom (M) is calculated using the Equation \ref{equ_loopwhite}.
\begin{equation}
M=11-6-3+1=3
\end{equation}
Therefore the mechanism has $3$ degrees of freedom.
\section{A comparison the Zebra Crossing Method with other quick methods}
\begin{sidewaysfigure}
\begin{center}
\includegraphics[scale=0.45]{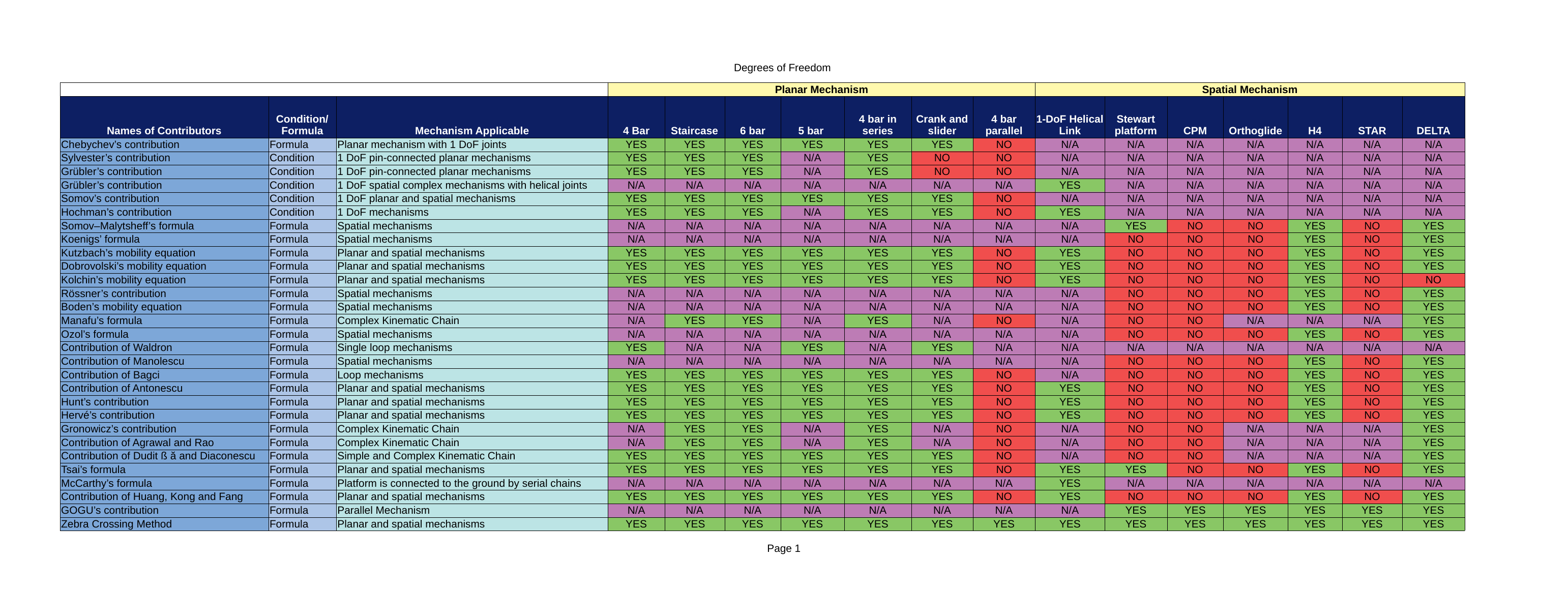} 
\caption{The zebra crossing algorithm compared with the other methods in the literature}
\label{fig_comparisionarchives}
\end{center}
\end{sidewaysfigure}

In \cite{gogu2005mobility}, details about the existing methods in the literature for calculating degrees of freedom of mechanisms are given. We compare the results of Zebra Crossing method with the quick methods mentioned in it for the mechanisms analysed above. The consolidated results are given in the Table \ref{fig_comparisionarchives}. It can be seen in the Table \ref{fig_comparisionarchives} that applying the quick methods for planar mechanisms gives the correct degrees of freedom for most of the cases. But considering the spatial mechanisms, especially the modern parallel manipulators, the quick methods fail to give the correct degrees of freedom. The reason for this is the inability to find kinematically dependant loops in the mechanisms. A detailed explanation for this is given in \citep{gogu2005mobility}. 

The proposed Zebra crossing method can be applied to both planar and spatial mechanisms. The Stewart platform, Cartesian Parallel manipulator, Orthoglide manipulator and the STAR manipulator are a few modern manipulators. Most of the quick methods that exist in the literature are either not applicable or give wrong results for these modern mechanisms. But when the Zebra crossing method is applied to these modern mechanisms, we obtain the correct degrees of freedom.

The traditional formulae used to calculate the degrees of freedom does not take into account the number of loops and the number of joints attached to the fixed link. The Zebra Crossing method takes into account these two critical parameters which aid in exact prediction of the degrees of freedom.

In summary, the Zebra Crossing method, proposed by us, can solve the degrees of freedom for all the given mechanisms. None of the other methods proposed in the literature have this capability. Even a combination  of all the earlier proposed methods are not be able to solve all the mechanisms given here which the Zebra Crossing method can do.  
 
\section{Conclusions}
A novel Zebra Crossing method to find the degrees of freedom for a mechanism is proposed in this work. The designed mechanism is converted into a zebra crossing diagram. The number of loops are calculated. Based on the number of loops and the nature of the mechanism, a suitable formula for calculating degrees of freedom is chosen. The degrees of freedom of the mechanism is calculated using it. To the best of authors knowledge, this quick method can be applied to any mechanism. About $14$ mechanisms of which a few of them whose degrees of freedom can not be found using the methods existing in the literature can be found using the zebra crossing method. Thus the proposed Zebra Crossing method is more universal in its applicability than other similar methods.  

\textbf{Acknowledgment:} 
The authors acknowledge partial funding support from RBCCPS, IISc Bangalore, Karnataka, India.

\section*{References}

\bibliography{mybibfile}
\end{singlespace}
\end{document}